\documentclass[letterpaper, 10 pt, conference]{ieeeconf}  % Comment this line out if you need a4paper

\IEEEoverridecommandlockouts                              % This command is only needed if 
\usepackage{graphics} % for pdf, bitmapped graphics files
\usepackage{epsfig} % for postscript graphics files
\usepackage{mathptmx} % assumes new font selection scheme installed
\usepackage{times} % assumes new font selection scheme installed
\usepackage{amsmath} % assumes amsmath package installed
\usepackage{amssymb}  % assumes amsmath package installed
\usepackage{kotex}
\usepackage{times}
\usepackage{epsfig}
\usepackage{makecell} % 추가해야 함
\usepackage{graphicx}
\usepackage{amsmath}
\usepackage{inconsolata}
\usepackage{amssymb}
\usepackage{pifont}
\usepackage{dblfloatfix}
\usepackage{url}
\usepackage[table,xcdraw]{xcolor} 
\usepackage{graphicx} 
\usepackage{caption}
\usepackage{microtype}
\usepackage{multirow}
\usepackage{float} 
\usepackage{placeins}
\usepackage{subfigure}
\usepackage{subcaption}
\usepackage{cite}
\usepackage{lipsum}

\definecolor{DarkYellow}{rgb}{0.87, 0.72, 0.0} % 원하는 RGB 값으로 정의
\definecolor{skyblue}{RGB}{0, 150, 255}  % 연한 하늘색
\definecolor{peach}{RGB}{255, 153, 102}    % 복숭아색 (PeachPuff)
\definecolor{lavender}{RGB}{148, 87, 235} % 진한 연보라색 (보라색 계열)
\definecolor{rose}{RGB}{255, 102, 102}
\definecolor{encoderpink}{RGB}{162,59,114}
\definecolor{decoderblue}{RGB}{46,134,171}
\definecolor{darkgreen}{RGB}{0,100,0}

\makeatletter
\renewcommand\@cite[2]{[\textcolor{skyblue}{#1}\if@tempswa , #2\fi]}
\makeatother

\usepackage{soul} %%% 형관펜 효과

\title{DroneKey\textcolor{encoderpink}{+}\textcolor{decoderblue}{+}: A Size Prior-free Method and New Benchmark \\ for Drone 3D Pose Estimation from Sequential Images
}
\author{Seo-Bin Hwang$^{1}$ and Yeong-Jun Cho$^{1*}$% <-this % stops a space
\thanks{$^{1}$Department of AI Convergence, Chonnam National University, Korea.
        {\tt\small cnu.cvl.hsb@gmail.com and \tt\small yj.cho@jnu.ac.kr}}%
\thanks{*Corresponding Author}% <-this % stops a space
}

\makeatletter
\let\@oldmaketitle\@maketitle % 기존 \maketitle 저장
\renewcommand{\@maketitle}{\@oldmaketitle% 기존 타이틀 출력 후 추가
  \begin{center}
  \vspace{15pt}
    \parbox{\linewidth}{%
      \centering
      \includegraphics[width=1\linewidth]{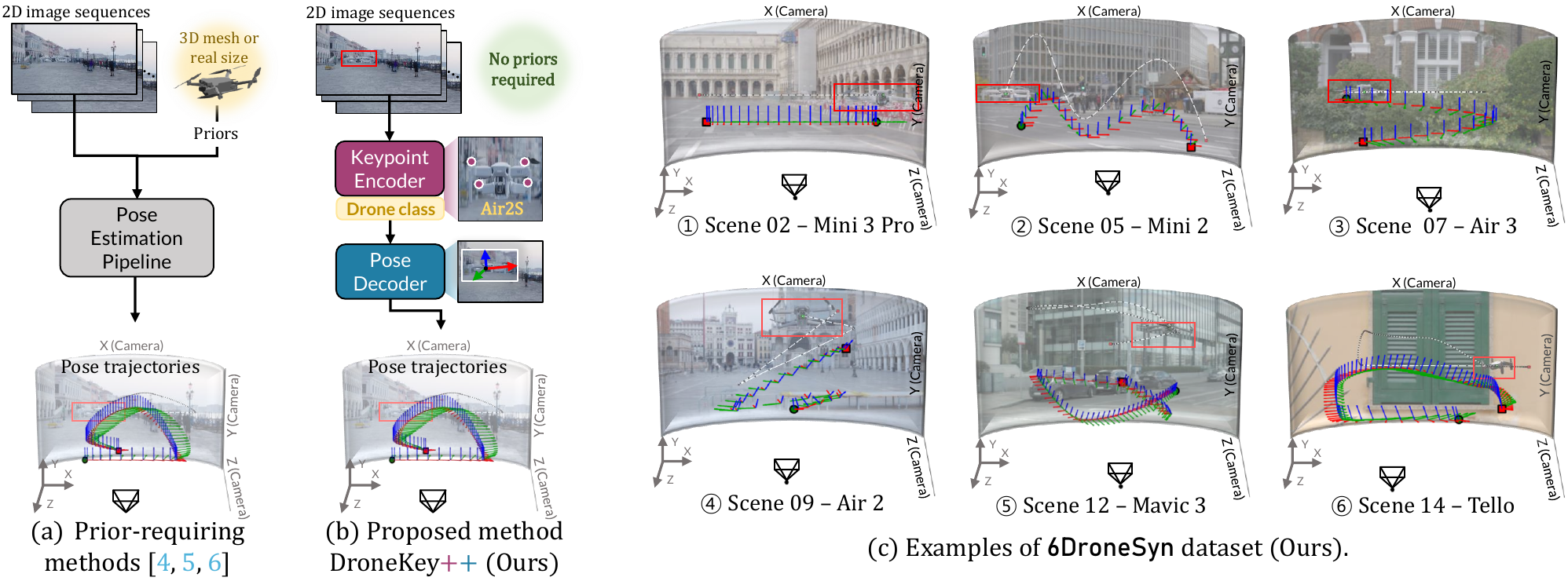}  % 원래는 7이었음
        \captionof{figure}{
            \textbf{Method overview and our new dataset.}
                (a) Existing silhouette-based methods are limited to single drone types and need 3D mesh.
                (b) Our proposed DroneKey\textcolor{encoderpink}{+}\textcolor{decoderblue}{+} integrates \textcolor{encoderpink}{keypoint encoder} and \textcolor{decoderblue}{3D pose decoder}, eliminating manual size input for various drone types.
                (c) Examples from our \texttt{6DroneSyn} dataset across diverse scenes and drone types with corresponding 3D pose trajectories.}
      \label{fig:01}
    }
  \end{center}
  \bigskip
  \vspace{-20pt}
}
\makeatother

\begin{document}
\maketitle
\thispagestyle{empty}
\pagestyle{empty}

\addtocounter{figure}{-1}
%■■■■■■■■■■■■■■■■■■■■■■■■■■■■■■■■■■■■■■■■■■■■■■■■■■■■■■■■■■■■■■■■■■■■■■■■■■■

\linespread{0.95} % 기본값은 1, 0.9로 줄이면 간격이 좁아짐

\begin{abstract}

    Accurate 3D pose estimation of drones is essential for security and surveillance systems.
    However, existing methods often rely on prior drone information such as physical sizes or 3D meshes. 
    At the same time, current datasets are small-scale, limited to single models, and collected under constrained environments, which makes reliable validation of generalization difficult.
    We present DroneKey\textcolor{encoderpink}{+}\textcolor{decoderblue}{+}, a prior-free framework that jointly performs keypoint detection, drone classification, and 3D pose estimation.
    The framework employs a keypoint encoder for simultaneous keypoint detection and classification, and a pose decoder that estimates 3D pose using ray-based geometric reasoning and class embeddings.
    To address dataset limitations, we construct \texttt{6DroneSyn}, a large-scale synthetic benchmark with over 50K images covering 7 drone models and 88 outdoor backgrounds, generated using 360-degree panoramic synthesis.
    %Feature distribution analysis demonstrates that our dataset achieves greater realism than existing synthetic datasets, effectively reducing the domain gap with real environments.
    Experiments show that DroneKey\textcolor{encoderpink}{+}\textcolor{decoderblue}{+} achieves MAE 17.34$^\circ$ and MedAE 17.1$^\circ$ for rotation, MAE 0.135 m and MedAE 0.242 m for translation, with inference speeds of 19.25 FPS (CPU) and 414.07 FPS (GPU), demonstrating both strong generalization across drone models and suitability for real-time applications.
% ■ (sb) 코드 공개
    The dataset is available at \textcolor{skyblue}{\url{will_be_able}}.
\end{abstract}

%■■■■■■■■■■■■■■■■■■■■■■■■■■■■■■■■■■■■■■■■■■■■■■■■■■■■■■■■■■■■■■■■■■■■■■■■■■■
%■■■■■■■■■■■■■■■■■■■■■■■■■■■■■■■■■■■■■■■■■■■■■■■■■■■■■■■■■■■■■■■■■■■■■■■■■■■
% \linespread{0.85} % 기본값은 1, 0.9로 줄이면 간격이 좁아짐
\section{INTRODUCTION}
\label{sec:introduction}

    \begin{table*}[t]
    \setlength{\tabcolsep}{4.5pt} % 기본은 6pt
    % \begin{table*}[b!]
        \centering
        \caption{\textbf{Comparison of drone datasets.} 
                Each dataset is compared across task type (Task), dataset name (Dataset), number of images (Frames), image resolution (Resolution), number of drone classes (Classes), number and type of backgrounds (Background, in=indoor/out=outdoor), inclusion of synthetic data (Syn), inclusion of real-world data (Real), number of sequences (Seq), annotation types (Annotation), and public availability (Published). N/A is not available.}
        \label{tab:01}
        \begin{tabular}{clccccccccc}
        \hline \noalign{\hrule height 0.5pt}
        \rowcolor[HTML]{FFFFFF} 
        \textbf{Task}                                                                                                 & \textbf{Dataset}    & \textbf{Frames} & \textbf{Resolution} & \textbf{Classes} & \textbf{Background} & \textbf{Syn} & \textbf{Real} & \textbf{Seq} & $\textbf{Annotation}^{*}$    & \textbf{Published} \\ \hline \noalign{\hrule height 0.5pt}
        \rowcolor[HTML]{FFFFFF} 
        \cellcolor[HTML]{FFFFFF}                                                                                      & \texttt{Drone Image Dataset}~\cite{drone-img-db} & 8,001         & Diverse             &  1        &     Diverse      &              & \checkmark    &     N/A    & 2D(B)                  & \checkmark       \\
        \rowcolor[HTML]{FFFFFF} 
        \cellcolor[HTML]{FFFFFF}                                                                                      & \texttt{Drone Dataset}~\cite{drone-db-UAV}       & 1,359         & Diverse             &   1       &     Diverse         &              & \checkmark    &  N/A      & 2D(B)                  & \checkmark       \\
        \rowcolor[HTML]{FFFFFF} 
        \multirow{-3}{*}{\cellcolor[HTML]{FFFFFF}\begin{tabular}[c]{@{}c@{}}2D Drone\\ Detection\end{tabular}}           & Hwang et al.~\cite{single-drone-traj}& 12,541        & Diverse             & 4              &    Diverse         & \checkmark   & \checkmark    &    N/A   & 2D(B)                  & \checkmark       \\ \hline
        \rowcolor[HTML]{FFFFFF} 
        \cellcolor[HTML]{FFFFFF}                                                                                      & Li. et al.~\cite{multi-drone-traj}    & N/A            & N/A                  & 1              & 2 (in+out)          &              & \checkmark    & 2              & 2D(B), 3D(t)           &                  \\
        \rowcolor[HTML]{FFFFFF} 
        \multirow{-2}{*}{\cellcolor[HTML]{FFFFFF}\begin{tabular}[c]{@{}c@{}}3D Trajectory\\ Reconstruction\end{tabular}} & Hwang et al.~\cite{single-drone-traj}      & 2,850         & 640x480             & 4              & 2 (out)             & \checkmark   & \checkmark    & 12             & 2D(B), 3D(t)           & \checkmark       \\ \hline
        \rowcolor[HTML]{FFFFFF} 
        \cellcolor[HTML]{FFFFFF}                                                                                      & Jin et al.~\cite{DronePoseGraph}           & 15,620        & N/A                   & 2              & 5 (in+out)          &              & \checkmark    & 6              & 2D(B), 3D(R+t)         &                  \\
        \rowcolor[HTML]{FFFFFF} 
        \cellcolor[HTML]{FFFFFF}                                                                                      & Albanis et al.~\cite{DronePose}                   & 55,988        & 320x240             & 1              & 90 (in)             & \checkmark   &               & 90             & 2D(M+B), 3D(R+t)       & \checkmark       \\
        \rowcolor[HTML]{FFFFFF} 
        \cellcolor[HTML]{FFFFFF}                                                                                      & You et al.~\cite{UAV-pose}                 & 27,351        & 1024x1024           & 4+             & 3 (out)             &              & \checkmark    & 3              & 2D(K+B), 3D(R+t)       &                  \\
        \rowcolor[HTML]{FFFFFF} 
        \cellcolor[HTML]{FFFFFF}                                                                                      & Hwang et al.~\cite{DroneKey}            & 11,200        & 1920x1080           & 2              & 4 (out)             & \checkmark   &               & 13             & 2D(K+B), 3D(R+t)       & \checkmark       \\
        \rowcolor[HTML]{EFEFEF} 
        \multirow{-5}{*}{\cellcolor[HTML]{FFFFFF}\begin{tabular}[c]{@{}c@{}}3D Pose\\ Estimation\end{tabular}}        & \textbf{\texttt{6DroneSyn} (Ours)}                                               & 52,920        & 1920x1080           & 7              &    88 (out)        & \checkmark   &      & 91             & 2D(K+B), 3D(R+t)       & Will be       \\ \hline \noalign{\hrule height 0.5pt}
        \end{tabular}
        \\[0.15cm]
        * In annotations, B=bounding box, K=keypoint, M=mask, R=rotation, and t=translation.
        \vspace{-15pt}
    \end{table*}

% ■ (sb) 문제 제기 (Problem Motivation)
    % □ (sb) 최근 불법적인 목적으로 드론을 오남용하는 사례가 급속히 증가하면서 국가 보안과 공공 안전에 심각한 위협이 되고 있다. 
    % □ (sb) 이러한 새로운 보안 위협에 대응하기 위해 무허가 드론 탐지 및 추적을 위한 안티드론 시스템의 필요성이 증대되고 있다[N].
    % □ (sb) 이러한 시스템의 핵심 구성 요소는 정확한 3차원 자세 추정으로, 
    % □ (sb) 이를 통해 드론의 시선 방향과 잠재적 표적 영역을 파악할 수 있어 드론 활동의 불법성과 위험도를 평가하는 데 중요한 정보를 제공한다.
    % (✅recently 추가) (✅ The key -> A key 변경)
    % □ (sb) 특히 실제 공공 안전 환경에서의 실용적 배치를 위해서는 단일 카메라 영상 기반 시스템 개발이 필수적이다.
    % □ (sb) 이는 다수의 센서나 특수 장비보다는 기존 인프라에 널리 보급된 CCTV와 같은 단일 카메라를 활용함으로써 달성할 수 있다.
    % □ (sb) 기존 드론 자세 추정 연구들은 주로 협력적 환경에서 내부 센서 접근 또는 다중 카메라 시스템을 가정한다.
    % ✅ (yjcho) "cooperative environments" 라는 말은 불필요 & 불명확 (제외)
    % □ (sb) 하지만 무허가 드론 탐지는 비협력적 환경으로, 내부 정보 접근이 불가능하고 광범위한 감시 영역에 다중 카메라 설치가 비현실적이다.
    % ✅ (yjcho) "non-cooperative environments 비협력적 환경"또한 불필요, 그냥 "접근이 불가능"라고만 설명해도 충분.
    % □ (sb) 따라서 기존 CCTV를 활용한 단일 카메라 기반 접근법이 현실적으로 유일한 해결책이다. ✅ viable 이런단어는 논문에서 잘 안쓰는것 같습니다.
        \textls[-1]{
        Recently, illegal drone use is rapidly increasing and threatens national security and public safety. 
        To address these threats, it is essential to develop anti-drone systems that can reliably detect and track unauthorized drones~\cite{anti-drone}.
        A key component of these systems is accurate 3D pose estimation, enabling the determination of a drone’s line of sight and potential target regions. 
        This provides critical information for assessing the threat level of drone activities.
        Existing drone 3D pose estimation studies mainly assume access to the drone's internal sensors~\cite{internal-senser} or rely on multi-camera settings~\cite{multi-drone-traj}.
        However, accessing drones’ internal sensor data is infeasible, and multi-camera setups are impractical in real-world settings.
        Given the widespread presence of CCTV in urban areas, single-camera approaches therefore represent the most practical and scalable solution for drone 3D pose estimation.
        }

% ■ (sb) 기존 연구의 방법론적 한계
    % □ (sb) 단일 카메라 영상으로부터 드론의 3차원 자세를 추정하기 위한 다양한 방법들이 연구되어 왔다.
    % □ (sb) 그러나 이러한 접근법들은 드론에 대한 사전 정보가 필요하다는 근본적인 한계로 인해 실제 적용에 어려움을 겪는다.
    % □ (sb) 첫째, 실루엣 기반 방법은 손실 함수에서 사용할 마스크 생성을 위해 3D 드론 메쉬가 필요하다. 
    % ✅ (yjcho) single drone에 대한 한계를 굳이 여기서 언급할 필요는 없을것 같고... related works나 exp에 작성할것-> 논지가 너무 여러개로 흩어지는것 같음. 또한 실루엣 기반의 방법이라고 명명하는것도 불필요. 실루엣 기반 방법보다는 3D mesh prior가 필수적인 방법이다라는것을 강조하는게 맥락상 좋음.
    % □ (sb) 게다가, 이 방법들은 단일 드론 모델에 대해서만 학습되기 때문에 다른 형태의 드론으로 일반화하기 어렵다.
    % □ (sb) 둘째, 키포인트 기반 방법은 PnP solver의 2D-3D 대응점 매칭 과정에서 발생하는 scale ambiguity로 인해 절대적인 크기 정보 없이는 정확한 거리 추정이 불가능하다.
    % □ (sb) 따라서 드론의 실제 크기 정보를 필수적으로 요구하며, 드론 모델별로 크기를 사전에 수동 입력해야 하는 제약이 있다.
    % □ (sb) 나아가, 최근 제안된 DroneKey는 드론 특화 키포인트 검출 방법을 제안했지만, 여전히 PnP 기반 파이프라인으로 인해 실제 드론 크기 정보 의존성 문제를 해결하지 못한다.
    % □ (sb) 이러한 한계들은 실제 환경에서 다양한 드론이 등장하는 상황에서 즉각적이고 확장성 있는 자세 추정을 어렵게 만든다.
        % \cite{dronemarker}\cite{DronePoseGraph}\cite{DronePose}\cite{DroneKey} 는 아래 처럼 수정. 추후 확인 필수
        Various methods~\cite{dronemarker}--\cite{DroneKey} have been proposed for drone 3D pose estimation using single-camera images.
        However, these approaches rely on prior information about drones, such as physical size or 3D shape priors, restricting their applicability in practice.
        For example, DronePose~\cite{DronePose} relies on 3D drone meshes prior for training, as shown in Fig.~\textcolor{skyblue}{\ref{fig:01}}a.
        Most drone pose estimation methods~\cite{DronePoseGraph}\cite{DroneKey} adopt keypoint detection with a PnP solver~\cite{pnp}. 
        However, these methods also rely on prior information about drones, such as physical size, which must be manually provided for each model and thus limits their applicability to unseen or diverse drones.

% ■ (sb) 그래서 우리가 제안하는 것 => ※ Ablation research를 통해 바뀌어야하는 부분!
    % □ (sb) 이러한 한계를 해결하기 위해 우리는 DroneKey++를 제안한다.
    \textls[-1]{
        To solve these problems, we propose DroneKey\textcolor{encoderpink}{+}\textcolor{decoderblue}{+} as illustrated in Fig.~\textcolor{skyblue}{\ref{fig:01}}b.
    % □ (sb) 이는 키포인트 검출, 자동 드론 분류, 그리고 3차원 자세 추정을 통합한 학습 프레임워크이다. 이를 통해 실제 크기나 3D 형상 등 드론 사전 정보 요구사항을 완전히 제거한다.
        This is a unified learning framework that integrates keypoint detection, drone classification, and 3D pose estimation. 
        It eliminates the need for prior drone information such as physical sizes or 3D meshes.
    % □ (sb) DroneKey++는 두 가지 핵심 모듈로 구성된다:
        DroneKey\textcolor{encoderpink}{+}\textcolor{decoderblue}{+} consists of two main components:
    % □ (sb) Keypoint Encoder는 DroneKey의 Transformer 구조에 class token을 추가하여 키포인트 검출과 드론 종류 분류를 동시에 수행한다.
        The keypoint encoder extends DroneKey's transformer~\cite{transformer} structure with class tokens~\cite{vit} to simultaneously perform keypoint detection and drone type classification.
    % □ (sb) 이를 통해 해당 클래스의 embedding을 생성하고 실제 크기 정보를 자동으로 인코딩한다.
        This enables the generation of class embeddings that automatically encode physical size information.
    % □ (sb) pose Decoder는 키포인트로부터 ray를 추정하고, drone의 class embedding feature와 결합하여 3D point를 추정한 후 이를 기반으로 드론의 3D pose를 예측한다.
        The pose decoder estimates rays from keypoints, combines them with class embedding features of the drone to estimate 3D points, and then predicts the drone's pose based on this.
    % □ (sb) 학습에는 키포인트와 3D pose 오차를 통합한 다중 손실 함수를 사용한다. 또한 훈련 과정에서 키포인트 검출에서 자세 추정으로 초점을 이동시키는 손실 가중치 스케줄링 전략을 적용한다.
        For training, we use a combined loss function that includes both keypoint and 3D pose errors. 
        % ✅ (yjcho) 아래 부분 제외
        % We apply a loss-weight scheduling strategy that shifts focus from keypoint detection to pose estimation during training.
    % □ (sb) 제안된 방법은 MAE X.X$^\circ$, RMSE X.X cm, XX FPS를 달성하며 기존 방법들을 능가한다. 
        Our method achieves MAE 17.34$^\circ$ (rotation), MAE 0.135m (translation), and 414.07 FPS (GPU, 19.25 FPS on CPU), outperforming existing approaches.
    }

% ✅ (yjcho) proposed methods 수정 이후 검토 예정.
% ■ (sb) 그런데 알고보니 dataset에도 문제가 있음. 그래서 dataset 구축도 필수적임. 그래서 만들었음!
    % ✅ 교수님: 앞 문단과 자연스럽게 이어지지 않음 
    % % □ (sb) 하지만 효과적인 학습과 검증을 위해서는 고품질 데이터셋이 필수적이다. 드론은 비교적 최근 객체로, 관련 연구들은 초기 단계에 머물러 있다.
    %     However, high-quality datasets are essential for effective training and validation. Drones are relatively recent objects, and related research remains in early stages.
    % \textls[-1]{
    % □ (sb) 이러한 방법론적 기여와 함께, 우리는 포괄적인 벤치마크 데이터셋도 구축하였다. 
    % □ (sb) 드론은 비교적 최근 객체로 관련 연구가 초기 단계에 있어, 기존 데이터셋들은 소규모이며 단일 드론 모델에 제한되어 있다.
    % □ (sb) 이는 일반화 성능과 방법론 검증의 신뢰성을 심각하게 제한한다.
    % □ (sb) 이러한 한계를 해결하기 위해 우리는 7가지 드론 모델과 88개의 다양한 실외 배경을 포함한 50K 이상의 대규모 synthetic dataset인 \texttt{6DroneSyn}을 구축했다.
    % □ (sb) 해당 데이터셋에는 3D pose, 2D 및 3D keypoint, 카메라 정보, bounding box 등이 포함되어있다. 
    % □ (sb) 이러한 포괄적 벤치마크는 기존 데이터셋 한계를 근본적으로 해결하고 실제 안티드론 시스템에 필요한 강건성을 보장한다.
    % □ (sb) 360도 파노라마 기반 합성 접근법을 통해 실제 환경과 유사한 고품질 데이터를 생성하였으며, Section V-D에서 feature distribution analysis를 통해 domain gap 문제를 효과적으로 해결했음을 증명한다.
        Along with these methodological contributions, we also constructed a large-scale benchmark dataset.
        Since drones are relatively recent objects of study, existing datasets are small-scale and often restricted to a single drone model, which severely limits generalization and the reliability of methodological validation.
        To overcome these limitations, we introduce \texttt{6DroneSyn}, a synthetic dataset with over 50K images covering 7 drone models and 88 diverse outdoor backgrounds (Fig.~\textcolor{skyblue}{\ref{fig:01}}c).
        The dataset provides rich annotations, including 2D and 3D keypoints, bounding boxes, camera information, and full 3D poses (rotation and translation).
        By addressing both scale and diversity, \texttt{6DroneSyn} enables robust evaluation of pose estimation methods for real-world anti-drone applications.
        Furthermore, through our 360-degree panorama-based synthesis approach, we generate realistic imagery, with feature distribution analysis in Sec.~\textcolor{skyblue}{\ref{sec:exp:dataset}} confirming effective domain gap reduction.
        % }

% ■ (sb) 제안 연그의 contributions
    % □ (sb) 본 연구의 주요 기여는 다음과 같다:
        The main contributions of this work are as follows:
        \begin{itemize}
            % \item \textbf{Prior-free framework.} 키포인트 검출, 자동 드론 분류, 그리고 3D 자세 추정을 통합하여 실제 크기나 3D 형상 등 드론 사전 정보 요구사항을 완전히 제거한다.
                \item \textbf{Prior-free framework.} \
                We integrate keypoint detection, drone classification, and 3D pose estimation into a unified framework, eliminating the need for prior drone information such as physical sizes or 3D meshes.
            % \item \textbf{새로운 벤치마크.} 실제 환경과의 도메인 갭을 최소화한 5만장 이상의 고품질 합성 데이터셋을 완전한 주석과 함께 제공한다.
                \item \textbf{New benchmark.} \
                We present \texttt{6DroneSyn}, a high-quality synthetic dataset with over 50K images and comprehensive annotations, specifically designed to minimize the domain gap with real-world environments.
                
            % \item \textbf{성능.} MAE X.X도 (회전), RMSE X.X cm, MAE X.X cm (위치), XX FPS를 달성하여 실시간 적용이 가능하다.
                \item  \textls[-1]{\textbf{Performance.} \ 
                Our method achieves MAE $17.34^\circ$ and MedAE $17.1^\circ$ for rotation, MAE $0.135$m and MedAE $0.242$m for translation, with inference speeds of 19.25 FPS (CPU) and 414.07 FPS (GPU), demonstrating suitability for real-time applications.
                }
        \end{itemize}
    % □ (sb) 이러한 기여를 통해 다양한 드론 모델에 대한 우수한 일반화 성능과 감시 응용에서의 강력한 잠재력을 가진 드론 자세 추정 시스템을 구현한다.
    % ✅ (yjcho) 이미 앞서 충분히 강조했으므로 생략
        %These contributions enable a drone pose estimation system with excellent generalization across drone models and potential for surveillance applications.

%■■■■■■■■■■■■■■■■■■■■■■■■■■■■■■■■■■■■■■■■■■■■■■■■■■■■■■■■■■■■■■■■■■■■■■■■■■■
%■■■■■■■■■■■■■■■■■■■■■■■■■■■■■■■■■■■■■■■■■■■■■■■■■■■■■■■■■■■■■■■■■■■■■■■■■■■

\begin{figure*}[t]
    \vspace{-10pt}
	\centering
	\includegraphics[width=1\linewidth]{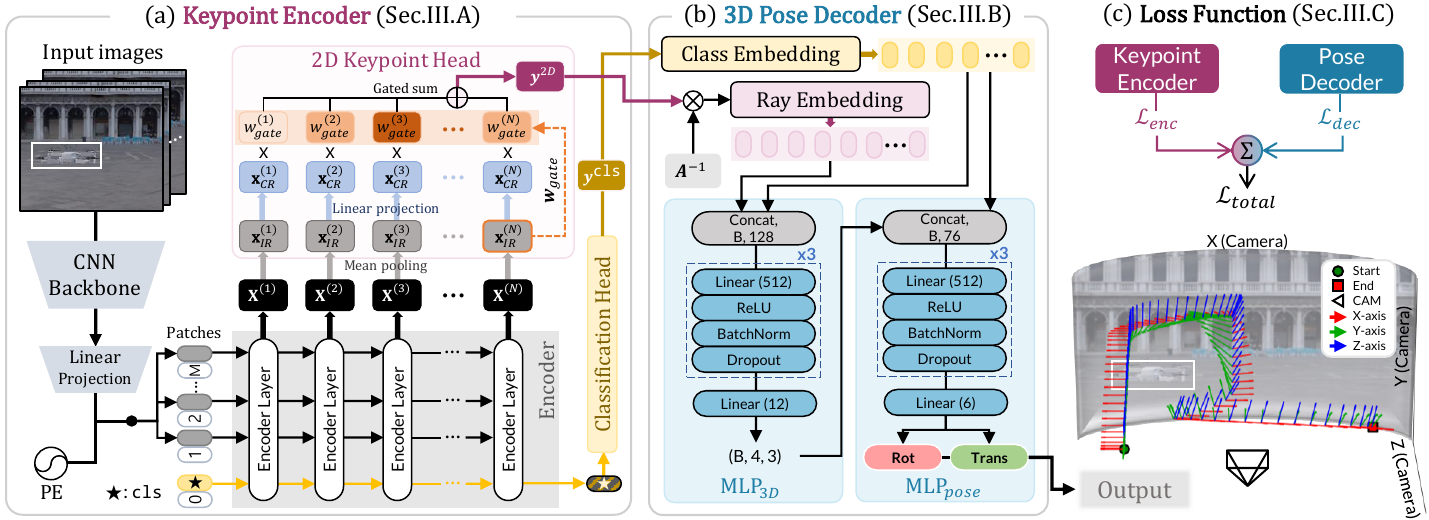}
	\caption{\textbf{Overall framework for DroneKey\textcolor{encoderpink}{+}\textcolor{decoderblue}{+}.} The proposed end-to-end pipeline consists of three main components: (a) keypoint encoder, which extracts 2D keypoints and drone class features from the input image sequence; (b) 3D pose decoder, which integrates class embedding and ray embedding to estimate the drone’s 3D rotation and translation; and (c) loss function, which combines encoder and decoder supervision into the total loss. This unified architecture enables simultaneous keypoint detection, class prediction, and accurate 3D pose estimation.}
	\label{fig:02}
    \vspace{-15pt}
\end{figure*}

% \vspace{10pt}
\section{RELATED WORKS}
\label{sec:relatedworks}

    \subsection{Drone Datasets}
    \label{sec:relatedworkd:dataset}
    % □ (sb) 드론은 다양한 분야에서 급속히 확산되고 있지만, COCO 같은 대규모 범용 데이터셋에는 포함되지 않았다\textcolor{skyblue}{\cite{coco}}.
        Despite rapid expansion across military, commercial, and civilian applications, drones are not included in large-scale general datasets like COCO~\cite{coco}.
    % □ (sb) 이에 따라 드론 관련 연구를 위한 전용 데이터셋들이 개발되기 시작했다. 
    % ✅ (yjcho) 이에따라->최근에서야 로 수정, 아래 문장과 병합
    % □ Table 1과 같이 드론 연구는 detection, trajectory reconstruction, 3D pose estimation의 세 영역으로 분류되며, 각각 특화된 데이터셋이 개발되었다.
        In recent years, specialized drone datasets have been developed for research areas such as drone detection, 3D trajectory reconstruction, and 3D pose estimation, as summarized in Tab.\textcolor{skyblue}{ ~\ref{tab:01}}.

        %As shown in Tab.\textcolor{skyblue}{\ref{tab:01}}, drone research is categorized into three main areas: detection, trajectory reconstruction, and 3D pose estimation, each with specialized datasets.

    %\noindent
    % □ (sb) \textbf{Detection} 영역에서는 초기에 Drone Image Dataset과 Drone Dataset이 제안되었다.
    % ✅ (yjcho) specialized datasets 이 계속 반복되는데, 앞에서 한번 specialized drone datasets 언급했으니, 뒤로부터는 그냥 drone datasets이라고 하면 됨 / 2D, 3D 인지 명확하게 명명.
    % □ (sb) 이들은 각각 8,001개, 1,359개의 영상으로 구성되어 기본적인 드론 검출 알고리즘 개발에 활용되었다.
    % □ (sb) 그러나 이들 데이터셋은 규모가 작아 성능 향상에 한계가 있었다.
        \noindent $\bullet$ \textbf{2D drone detection.} \quad It was the first area to develop drone datasets, with early works including \texttt{Drone Image Dataset}~\cite{drone-img-db} and \texttt{Drone Dataset}~\cite{drone-db-UAV}.
        These datasets contain 8,001 and 1,359 images, respectively, and have been primarily used for training 2D drone detection models.
        % ✅ (yjcho) 제외 However, these datasets had limited scale, restricting performance improvements.
        In addition, \texttt{2Drone (on+Aug)}~\cite{single-drone-traj} was introduced, generating 12,541 large-scale training samples by synthesizing 4 DJI drone models with diverse backgrounds through data augmentation.
    
    %\noindent
    % □ (sb) \textbf{Trajectory Reconstruction} 영역에서는 2D 영역 정보만 제공하는 Detection 대비 3차원 정보를 포함한 데이터셋이 구축되었다.
        \noindent $\bullet$ \textbf{3D trajectory reconstruction.} \quad Unlike detection datasets that provide only 2D annotations, 3D trajectory reconstruction datasets include both 2D images and 3D drone positions for each frame.
        %developed datasets with 3D information, compared to detection, which only provides 2D region information.
    % □ (sb) Li et al.은 multi-camera 환경에서 촬영된 샘플을 포함한 데이터셋을 제안했지만 비공개 정책으로 인해 연구 접근성이 제한적이다.
        Li et al.~\cite{multi-drone-traj} proposed a dataset captured in multi-camera environments, but it is not publicly available.
    % □ (sb) Hwang et al.은 4개 UAV 모델과 12개 시나리오를 기반으로 2,850개의 공개 샘플을 제공하는 Syn3Drone을 제안하여 궤적 추정 및 재구성 연구를 지원한다.
        Hwang et al.~\cite{single-drone-traj} proposed \texttt{Syn3Drone}, which provides 2,850 drone frames based on 4+ drone models and 12 scenarios, supporting trajectory estimation and reconstruction research.
    % □ (sb) 그러나 궤적 데이터는 드론의 위치 정보(translation)만 제공하고 회전 정보(rotation)는 포함하지 않아, 드론의 시선 방향이나 자세 분석이 중요한 보안 응용에서는 활용이 제한된다.
        However, these data only provide drone positions without rotation, limiting their use in security applications where orientation and viewing direction are critical.

    % \noindent
    % □ (sb) \textbf{3D Pose Estimation}은 counter-UAV 시스템에서 가장 중요한 기술적 요소로, 드론의 시선 방향과 위협도 분석에 필수적이다. 
        \noindent $\bullet$ \textbf{3D pose estimation.} 
        %\quad It is the most critical technology for anti-drone systems, essential for analyzing drone viewing direction and threat assessment. 
    % □ (sb) 이 영역에서는 UAVA, UAV-ADT, 3DKey+3DPose 등의 데이터셋이 제안되었으나 모두 실용적 한계를 보인다.
        \quad Several datasets, including \texttt{UAVA}~\cite{DronePose}, \texttt{UAV-ADT}~\cite{UAV-pose}, and \texttt{3DKey+3DPose}~\cite{DroneKey} have been proposed for drone 3D pose estimation.
    % □ (sb) Albanis et al.의 UAVA는 55,988개의 상당한 규모를 가지지만 단일 UAV 모델과 저해상도(320×240)로 제한되며, You et al.의 UAV-ADT는 실제 데이터를 포함하지만 27,351개의 제한된 규모를 가진다.
        \texttt{UAVA}~\cite{DronePose} has substantial scale (55,988 images) but is limited to a single drone model and low resolution (320$\times$240), while \texttt{UAV-ADT}~\cite{UAV-pose} includes real data but has limited scale (27,351 images).
    % □ (sb) Hwang et al.의 3DKey+3DPose는 고해상도 데이터를 제공하지만 2개 UAV 모델로만 제한되어 있다. 
        \texttt{3DKey+3DPose}~\cite{DroneKey} provides high-resolution data but is restricted to only 2 drone models. 
    % □ (sb) 결과적으로 모든 기존 데이터셋이 드론 모델 다양성과 실환경 일반화 측면에서 한계를 보인다.
        Consequently, all existing datasets have limitations in drone model diversity and real-world generalization.

    % ■ (sb) 한계점 + 어떤 db가 필요해?
    % □ (sb) 실제 환경에서 드론의 정밀한 3D 자세 정보를 수집하는 것은 거의 불가능하기 때문에, 대부분의 연구에서 합성 데이터를 활용한다.
        Since collecting precise 3D drone pose information in real-world environments is nearly impossible, most research relies on synthetic data.  
    % □ (sb) 실제로 공개된 드론 pose estimation 데이터셋들은 모두 합성 데이터이며, 이는 실제 환경에서의 정확한 ground truth 획득이 극도로 어렵기 때문이다.
        In fact, all publicly available drone pose estimation datasets are synthetic, as obtaining accurate ground truth in real environments is extremely challenging due to safety regulations, equipment limitations, and cost constraints.
    % □ (sb) 그러나 기존 합성 데이터들은 실제 환경과의 차이로 인한 domain gap 문제와 드론 모델 다양성 부족으로 한계를 보인다.
        However, existing synthetic data suffer from domain gap issues due to differences from real environments and lack of drone model diversity.
    % □ (sb) 따라서 실제 환경과 유사한 합성 데이터 생성과 다양한 드론 모델을 포괄하는 종합적 벤치마크 구축이 필수적이다.
        Therefore, generating synthetic data similar to real environments and building comprehensive benchmarks covering diverse drone models is essential.
    % % □ (sb) 이에 따라 본 연구에서는 7개 드론 모델과 88개 실외 배경을 포함한 대규모 합성 데이터셋 \texttt{6DroneSyn}을 구축하여 기존 데이터셋의 한계를 해결하고자 한다.
        % Accordingly, we construct \texttt{6DroneSyn}, a large-scale synthetic dataset with 7 drone models and 88 outdoor backgrounds, to address the limitations of existing datasets.

    \subsection{Drone 3D Pose Estimation Methods}
    \label{sec:relatedworks:drone3d}
    % ✅ 교수님: 앞서 해결하고자 하는 문제는 'prior'이기 때문에, 여기서도 그 내용에 관해 중점적으로 이야기해야함. 예를 들어, 실루엣 기반 -> 3d mesh, keypoint -> real size 형식으로 변경!
    % □ (sb) 드론 3D 자세 추정 연구는 안티드론 시스템과 감시 응용에서 중요성이 증대되고 있지만 상대적으로 제한적인 연구가 이루어지고 있다.
        Drone 3D pose estimation is becoming increasingly important for anti-drone systems and surveillance, yet it remains underexplored.
    % ㅁ (sb) 기존 연구를 '필요한 사전 정보(prior)'의 관점에서 정리
        %Existing studies can largely be divided according to the type of prior information they require: 
        %methods that depend on 3D mesh prior and methods that depend on physical size prior.
        Existing studies can mainly be categorized into two types: those that depend on 3D mesh priors and those that depend on physical size priors.
    % ㅁ (sb) 3D mesh priors: 마스크/실루엣 손실 계산을 위해 메쉬 기반 렌더링 필요. DronePose 예시.
    % ✅ (yjcho) 아래문장 논리가 이상해요. A 3D mesh prior가 실루엣 마스크 렌더링을 위해 필요하다-> 예전에 실루엣-based 이렇게 써놓은게 남아있는것 같은데. 분류 체계가 이상합니다. 또 synthetic--real domain gap 을 여기서 이야기하는게 맞나 싶네요. 데이터셋 이야기 아닌가요? 일단 그대로 둘게요.
        %\noindent $\bullet$ \textbf{A 3D mesh prior} is required to render silhouette masks for loss computation. 
        For example, DronePose~\cite{DronePose} adopts synthetic training with a silhouette loss, which requires drone 3D mesh priors to render silhouette masks. 
    % ✅ (yjcho) 앞에서 내내 prior 이야기를 하다가 여기서는 왜또, single drone 적용 한계나 synthetic--real domain gap이라는 다른 한계점을 지적하시나요?
    %    However, its reliance on a single drone model and the synthetic--real domain gap hinder generalization to drones of different shapes.
        However, its reliance on a single drone model and 3D mesh priors hinders generalization to drones of different shapes.
    % ㅁ (sb) Absolute size priors: PnP의 스케일 모호성 때문에 반드시 실제 크기 정보가 필요. Graph network, DroneKey 예시.
        %\noindent $\bullet$ \textbf{A physical size prior} is necessary because PnP solvers\textcolor{skyblue}{\cite{pnp}} inherently suffer from scale ambiguity in 2D--3D correspondence. 
        Jin et al.~\cite{DronePoseGraph} leveraged relational graph networks to enhance accuracy, while DroneKey~\cite{DroneKey} proposed gated key-representations. Nevertheless, both approaches depend on prior knowledge of the drone’s physical size to establish 2D--3D correspondences via PnP solvers~\cite{pnp}.
    
    % ㅁ (sb) 두 접근법 모두 '사전 정보 의존성'이라는 공통 한계가 있으며, 우리가 해결해야 할 문제임을 강조
    \textls[-1]{
        In summary, both 3D mesh priors and the size priors restrict scalability and prevent immediate deployment to unknown drones in practical environments. 
        This highlights the need for \textit{prior-free} methods that can automatically learn class and scale cues directly from images. Such methods would enable robust 3D pose estimation across diverse and unseen drone types.
    }

%■■■■■■■■■■■■■■■■■■■■■■■■■■■■■■■■■■■■■■■■■■■■■■■■■■■■■■■■■■■■■■■■■■■■■■■■■■■
%■■■■■■■■■■■■■■■■■■■■■■■■■■■■■■■■■■■■■■■■■■■■■■■■■■■■■■■■■■■■■■■■■■■■■■■■■■■
\section{PROPOSED METHODS} 
\label{sec:methods} 
% ■ (sb) architecture overview
    % □ (sb) Fig.\textcolor{skyblue}{\ref{fig:02}}에 나타난 바와 같이, DroneKey++는 키포인트 검출, 드론 분류, 3D pose 자세 추정을 하나의 통합 구조로 설계하였다.
        As shown in Fig.~\textcolor{skyblue}{\ref{fig:02}}, we propose DroneKey\textcolor{encoderpink}{+}\textcolor{decoderblue}{+}, an end-to-end framework that integrates keypoint detection, drone classification, and 3D pose estimation.
        It consists of two main networks: a \textcolor{encoderpink}{\textbf{keypoint encoder}} and a \textcolor{decoderblue}{\textbf{3D pose decoder}}.
    % □ (sb) Keypoint Encoder는 입력 이미지로부터 2D 키포인트와 드론 클래스를 동시에 추정하고, Pose Decoder는 이를 활용하여 3D pose를 직접 회귀한다. => ※ 두 문장으로 쪼개
        The keypoint encoder detects 2D keypoints and the drone’s class from input images. 
        The 3D pose decoder then estimates the drone’s 3D pose, including translation and rotation, based on the extracted keypoints and class features.
    % □ (sb) 훈련 과정에서 모든 작업의 손실을 결합한 손실 함수를 사용한다. 
        % The training process uses a combined loss function for all tasks.
    % □ (sb) 이러한 통합 구조는 기존 방법들의 사전 정보 의존성을 제거하고 다양한 드론 모델에 대한 일반화 성능을 제공한다. => ※ 너무 어려움 표현들이. 쉽고 짧은 문장으로 말해바 
        %This approach eliminates the need for prior drone knowledge, such as 3D mesh and physical size.
        %It can handle different types of drones effectively.
    % ✅ (yjcho) 위에 3문장 제외 -> 아래 문장 추가
        Unlike existing approaches that rely on prior information such as the drone’s physical size, our method is designed to operate in a prior-free manner. Specifically, the network first infers the drone’s class  from the input images, and this inferred class is then exploited within the decoder network to guide the learning of 3D pose representations. 
        Through this design, the network can estimate the drone’s full 3D pose directly from images alone, without requiring any external priors.

    %□□□□□□□□□□□□□□□□□□□□□□□□□□□□□□□□□□□□
    \subsection{Keypoint Encoder} % 색 제외
    \label{sec:methods:keypointencoder}
    % ✅ (yjcho) 아래 두문장 한문장으로 clear하게 정리.
    % ㅁ (sb) 드론 자세 추정을 위해서는 객체의 기하학적 구조를 대표할 수 있는 시각적 특징점인 키포인트가 필요하다. 
        %For drone 3D pose estimation, keypoints are required as visual feature points that can represent the geometric structure of the object. 
    % ㅁ (sb) 드론의 경우 프로펠러가 모든 드론 모델에 공통으로 존재하면서 기하학적 구조를 효과적으로 나타내는 시각적 특징을 가지므로 키포인트로 선정한다. 
        %In the case of drones, propellers are generally selected as keypoints since they exist commonly across all drone models and possess visual characteristics that effectively represent geometric structure. 
        For 3D drone pose estimation, keypoints are essential to capture the object’s geometric structure. Propellers are commonly chosen as keypoints because they are present across most drone models and provide  geometric cues.
    % ㅁ (sb) 그러나 프로펠러들 간의 높은 시각적 유사성으로 인해 기존 CNN 기반 방법들은 키포인트 간의 전역적 관계를 파악하는데 한계를 보인다. 
        %However, traditional CNN-based methods\textcolor{skyblue}{\cite{MaskRCNN, YOLO}} face limitations in capturing global relationships between keypoints due to the high visual similarity among propellers.
        However, traditional CNN-based methods~\cite{MaskRCNN}\cite{YOLO} face challenges in reliable keypoint detection, since propellers appear highly similar across a drone and also require consistent ordering to recover the 3D geometry.
        To address this issue, DroneKey~\cite{DroneKey} introduced a specialized method for extracting drone keypoints.
        %⚠️ Inspired by this work, we adopt DroneKey~\cite{DroneKey}’s transformer-based encoder~\cite{transformer}, which leverages self-attention to learn spatial arrangements among keypoints. => 저자 동일성을 암시하는 뉘앙스 제거
        To address this issue, prior work such as DroneKey~\cite{DroneKey} proposed a transformer-based encoder~\cite{transformer}, which leverages self-attention to learn spatial arrangements among keypoints.
    % ㅁ (sb) 이러한 문제를 해결하기 위해 공간 배치를 학습할 수 있는 self-attention 메커니즘을 가진 DroneKey\textcolor{skyblue}{\cite{DroneKey}}의 Transformer\textcolor{skyblue}{\cite{transformer}} encoder 구조를 활용한다. 
        %To address this limitation, we utilize DroneKey\textcolor{skyblue}{\cite{DroneKey}}'s Transformer\textcolor{skyblue}{\cite{transformer}} encoder with self-attention mechanisms capable of learning spatial arrangements.  
    % ㅁ (sb) 여기에 ViT\textcolor{skyblue}{\cite{vit}}의 CLS 토큰을 결합하여 사전 정보 없이 드론 분류와 크기 정보를 자동으로 학습하는 키포인트 인코더를 제안한다.
        %We propose a keypoint encoder that combines vision transformer (ViT)\textcolor{skyblue}{\cite{vit}}'s class (CLS) token with this architecture to automatically learn drone classification and size information without prior knowledge.
        % ✅ (yjcho) 아래와 같이 수정
        
        Unlike prior methods that rely on explicit priors such as the drone’s physical size for 2D--3D matching, we extend the DroneKey~\cite{DroneKey} keypoint encoder with a classification head.
        By leveraging the ViT~\cite{vit} [\texttt{cls}] token, the encoder predicts drone class information, guiding the learning of 3D geometry for prior-free pose estimation.
    % ㅁ (sb) 제안된 인코더는 공간적 키포인트 추출과 의미적 드론 분류를 동시에 수행한다.
        As a result, the proposed encoder jointly performs spatial keypoint extraction and drone classification.
    % □ (sb) 이를 위해 먼저, Transformer에 입력하기 위한 특징맵 $\mathbf{X}$를 추출하기 위해 입력 이미지로부터 CNN backbone인 ResNet을 사용한다.
    
        To achieve this, we employ ResNet~\cite{resnet} as the CNN backbone to extract feature maps $\mathbf{X}$ from input images for transformer input.
    % □ (sb) 이 특징맵을 패치 단위로 토큰화하여 임베딩으로 변환하고, positional encoding을 추가하여 $\mathbf{X}^{(0)}$로 표기하며, 이는 공간적 위치 정보가 포함된 초기 입력 표현이다.
        The feature map is then tokenized into patches, converted to embeddings, and combined with positional encoding to form $\mathbf{X}^{(0)}$, which represents the initial input with spatial position.
    % □ (sb)여기서 $\mathbf{x}_i^{(0)}$는 $i$번째 토큰을 나타내며, 전체 집합 $\mathbf{X}^{(0)} = \{\mathbf{x}_1^{(0)}, \dots, \mathbf{x}_M^{(0)}\}$으로 구성된다.
        $\mathbf{x}_i^{(0)} \in \mathbb{R}^d$ represents the $i$-th token of dimension $d$, and the complete set is composed of $\mathbf{X}^{(0)} = \{\mathbf{x}_1^{(0)}, \dots, \mathbf{x}_M^{(0)}\} \in \mathbb{R}^{d\times M}$.
    % □ (sb) 이렇게 변환된 토큰 특징들은 N개의 self-attention layer에서 순차적으로 처리된다:
        A learnable [\texttt{cls}] token and the token set are jointly fed into $N$ self-attention layers:
        \begin{equation}
            \left[\textbf{x}^{(l)}_{\texttt{cls}}, \ \mathbf{X}^{(l)}\right] = \text{selfAttn}\left(Q, K, V = \left[\textbf{x}^{(l-1)}_{\texttt{cls}}, \ \mathbf{X}^{(l-1)}\right]\right),
        \end{equation}
    % □ (sb) 여기서 $l = 1, 2, ..., N$은 transformer encoder layer의 인덱스를 나타내고, $\mathbf{X}^{(l)}$은 $l$번째 layer를 통과한 후의 전체 토큰 특징 표현이다. 
        where, $l = 1, 2, ..., N$ represents the index of transformer encoder layers and $\mathbf{X}^{(l)}$ is the set of feature representations after passing through the $l$-th layer.
    % □ (sb) Self-attention 연산을 위한 Q, K, V는 각각 query, key, value를 의미한다.
        $Q$, $K$ and $V$ for self-attention operations, represent query, key, and value.

    % □ (sb) 각 encoder layer에서 나온 특징들을 성공적으로 통합하기 위해 DroneKey에서는 gated summation을 활용한다. 
        To effectively integrate features across encoder layers, we adopt a gated summation mechanism, following DroneKey~\cite{DroneKey}.
    % □ (sb) 키포인트 검출을 위해 최종 layer 출력만 사용하는 대신, 모든 중간 표현들 $\{\mathbf{X}_{EN}^{(1)}, ..., \mathbf{X}_{EN}^{(N)}\}$을 함께 고려한다. 
        %Instead of using only the final layer output for keypoint detection, 
        %All intermediate representations $\{\mathbf{X}^{(1)}, ..., \mathbf{X}^{(N)}\}$ are jointly considered.
        % ✅ (yjcho) 아래 논리 전개가.. 이랬다 저랬다 (지금 문단은 “선형 변환 필요/언급 → mean pooling → 다시 선형 변환” 이런 식으로 왔다갔다)
        % ✅ (yjcho) 중복된 부분은 줄이거나, 독자가 헷갈리지 않도록 순서대로 설명할것.
    % □ (sb) 그러나 이러한 intermediate feature들은 크기와 표현 방식이 서로 다르므로, 선형 변환을 통해 키포인트 차원으로 축소한 뒤 gated summation을 적용한다.
    % □ (sb) 이를 위해 먼저 각 layer에 mean pooling을 적용하여 intermediate representation $\mathbf{X}_{IR}^{(l)}$을 생성한다. 
    % □ (sb) 이후 linear projection으로 compact representation $\mathbf{X}_{CR}^{(l)} = \mathbf{W} \cdot \mathbf{X}_{IR}^{(l)} + \mathbf{b}$를 만든다. 
        % Since these features differ in dimensionality, we apply a linear transformation to project them into the keypoint space before performing gated summation.
        % For this purpose, mean pooling is first applied to each layer to generate intermediate representation $\mathbf{X}_{IR}^{(l)}$.
        % Subsequently, linear projection creates compact representation $\mathbf{X}_{CR}^{(l)} = \mathbf{W} \cdot \mathbf{X}_{IR}^{(l)} + \mathbf{b}$.
        %Since features from different layers vary in dimensionality, 
        %Each intermediate representation $\mathbf{X}^{(l)}\in \mathbb{R}^{d\times N}$.
        %We apply mean pooling to each layer to obtain an intermediate representation $\mathbf{X}_{IR}^{(l)}$.
        %For each layer, the feature representation is given by $\mathbf{X}^{(l)} \in \mathbb{R}^{d \times M}$.
        For the output of each layer $\mathbf{X}^{(l)}$, we perform max pooling, yielding an intermediate representation $\mathbf{X}_{IR}^{(l)} \in \mathbb{R}^{d}$.
        We then apply a linear projection to map $\mathbf{X}_{IR}^{(l)}$ into the keypoint space,
        \begin{equation}
        \mathbf{X}_{CR}^{(l)} = \mathbf{W} \cdot \mathbf{X}_{IR}^{(l)} \in \mathbb{R}^{4\times2},    
        \end{equation}
        where of the four rows corresponds to four keypoints and the two columns represent its $(x,y)$ coordinates.
        The resulting compact representations $\mathbf{X}_{CR}^{(l)}$ are subsequently integrated across layers through gated summation.
         %to generate compact representations in the keypoint space, which are subsequently integrated through gated summation.      
    % □ (sb) 최종 layer의 중간 표현에서 생성된 가중치 $\mathbf{w}_{gate} = \text{softmax}(\mathbf{W}_g \cdot \mathbf{X}_{IR}^{(N)} + \mathbf{b}_g)$를 사용한 gated summation으로 모든 layer의 특징을 효과적으로 집약한다. 최종 키포인트 좌표는 다음과 같이 예측된다:
        To this end, we learn a weight vector of dimension $N$ from the final layer’s intermediate representation $\mathbf{X}_{IR}^{(N)}$ as follows:
        \begin{equation}
            \mathbf{w}_{gate} = \text{softmax}\left(\mathbf{W}_g \cdot \mathbf{X}_{IR}^{(N)}\right)=\left[w^{(1)}_{gate},w^{(2)}_{gate},\ldots,w^{(N)}_{gate} \right],    
        \end{equation}
        where $\mathbf{W}_g\in \mathbb{R}^{N\times d}$ is a projection matrix.
        %generated from the final layer's intermediate representation. 
        Final keypoint coordinates are predicted as follows:
        \begin{equation}
            \mathbf{y}^{2D} = \text{ReLU}\left(\sum_{l=1}^{N}w_{gate}^{(l)}\mathbf{X}_{CR}^{(l)}\right).
        \end{equation}
    % □ (sb) ReLU 함수는 비현실적인 음수 예측을 방지하고 안정적인 키포인트 표현 학습을 가능하게 한다.
        The Rectified Linear Unit~(ReLU) prevents unrealistic negative predictions and enables stable representation learning.

    % ✅ (yjcho) 기존에는 CLS 토큰에 대해 완전 따로 설명하고 있으며, 수식으로 표현도 안되는 상태. 수정된 내용에서는 수식 (1)에서 CLS토큰을 함께 설명했기 때문에 여기서 다시 설명할 필요가 없음.
    % □ (sb) 드론 클래스를 키포인트 검출과 동시에 추정하기 위해 learnable한 CLS token을 시각적 토큰들과 함께 입력에 추가한다. 
        %To estimate drone classes simultaneously with keypoint detection, a learnable CLS token is added to the input along with visual tokens.
        % □ (sb) CLS token $\mathbf{x}_{cls}^{(0)}$은 학습 과정에서 전체 이미지의 글로벌 정보를 집약하여 드론 타입을 분류하는 역할을 한다. 
    % □ (sb) 앞서 키포인트 검출을 위해 처리된 self-attention 과정을 CLS token도 함께 거치게 된다. 
        %The CLS token also undergoes the same self-attention process previously used for keypoint detection.
    % □ (sb) 그 결과 마지막 layer를 통과한 CLS token인 $\mathbf{x}_{cls}^{(N)}$은 별도의 classification head로 전달되어 다음과 같이 드론 클래스를 예측한다:
    % □ (sb) 여기서 $\mathbf{W}_{cls}$는 분류를 위한 선형 변환 행렬이고, $\mathbf{y}^{class}$는 드론 클래스에 대한 확률 분포를 나타낸다.
        To estimate drone classes alongside keypoint detection, we use the final [\texttt{cls}] token output $\mathbf{x}^{(L)}_\texttt{cls}$, which encodes global image information.
        This representation is then passed to a separate classification head to predict drone types as follows:
        \begin{equation}
            \mathbf{y}^\texttt{cls} = \text{softmax}\left(\mathbf{W}_\texttt{cls} \cdot \mathbf{x}_\texttt{cls}^{(N)}\right),
        \end{equation}
        where, $\mathbf{W}_\texttt{cls}$ is the linear transformation matrix for classification and $\mathbf{y}^\texttt{cls}$ represents the probability distribution over drone classes.

    %□□□□□□□□□□□□□□□□□□□□□□□□□□□□□□□□□□□
    
    \subsection{3D Pose Decoder}
    \label{sec:methods:posedecoder}
        Traditional approaches compute drone 3D pose by combining keypoints with explicit priors, e.g., the drone’s physical size.
        In contrast, we propose a novel decoder that allows the network to learn this estimation process directly, without relying on priors.
        By leveraging the class probability distribution $\mathbf{y}^\texttt{cls}$ from the encoder, the decoder learns drone class-specific 3D pose estimation that is more robust to errors than conventional PnP-based~\cite{pnp} 2D--3D matching.
    % □ (sb) 전체 과정은 클래스 임베딩, ray 추출, 특징 융합, 3D 키포인트 예측, 그리고 최종 3D pose 회귀로 구성된다.  
        The overall process consists of class embedding, ray extraction, feature fusion, and prediction of 3D keypoints and pose.

    % □ (sb) 먼저 키포인트 인코더로부터 예측된 2D 키포인트와 카메라 내재 파라미터로부터 정규화된 ray direction vector를 계산한다:
        First, we compute normalized ray direction vectors from predicted 2D keypoints and camera intrinsic parameters by
        \begin{equation}
           \mathbf{v}_k = \mathbf{A}^{-1}[\mathbf{y}^{2D}_{k}, 1]^{\top},
        \end{equation}
    % □ (sb) 여기서 $\mathbf{A}$는 카메라 내재 행렬, $\mathbf{y}^{2D}_i$는 $i$번째 키포인트의 예측된 2D 좌표, $\mathbf{v}_i$는 해당 키포인트에 대응하는 ray direction이다.  ✅ (교수님) A의 dimension 추가
        where $\mathbf{A} \in \mathbb{R}^{3\times 3}$ is the camera intrinsic matrix, $\mathbf{y}^{2D}_k$ is the predicted 2D coordinate of the $k$-th keypoint, and $\mathbf{v}_k$ is the corresponding ray direction.
    % ↔ (sb) Ray direction \mathbf{v}를 임베딩하여 단순 좌표 대신 네트워크가 처리 가능한 표현 공간인 e_{ray}로 변환한다.
        The computed ray direction $\mathbf{v}$ is embedded into $\mathbf{e}_{ray} \in \mathbb{R}^{64}$, transforming raw coordinates into a structured feature space that can be effectively exploited by the network.
    % ↔ (sb) 키포인트 인코더에서 예측된 드론 클래스는 $\mathbf{e}_{\texttt{cls}} \in \mathbb{R}^{64}$로 임베딩 되며, 이 표현은 드론의 물리적 크기에 관한 단서를 제공한다.
    % ↔ (sb) Ray direction embedding $\mathbf{e}_{ray}$와 클래스 임베딩 $\mathbf{e}_{cls}$를 concatenation으로 결합하여 융합 특징 $\mathbf{f}_{fused}$를 구성한다.
        The drone class predicted by the keypoint encoder is also embedded into $\mathbf{e}_{\texttt{cls}} \in \mathbb{R}^{64}$, which provides cues about the drone's physical size. 
        For these embeddings, we learn simple linear projection layers and then concatenate $\mathbf{e}_{ray}$ and $\mathbf{e}_{\texttt{cls}}$ to form the fused feature $\mathbf{f}_{fused}$.

    % □ (sb) 이후 MLP로 구성된 3D keypoint regression head가 융합 특징으로부터 드론의 상대적 3D 키포인트 좌표를 추정한다.

    % ✅ (yjcho) 여기 아래서부터 수정중.
        Next, an MLP-based 3D keypoint head estimates the relative 3D keypoint coordinates of the drone from the fused features as follows:
        \begin{equation}
            \mathbf{y}^{3D} = \text{MLP}_{3D}\left(\mathbf{f}_{fused}\right).
        \end{equation}
        It reconstructs 3D positions from 2D observations and learned class information.
    % □ (sb) 네트워크는 드론 키포인트의 3차원 상대 위치를 제공한다. ✅ (yjcho) 아래 내용 불필요 삭제
        %The network provides the 3D relative positions of drone keypoints.
    % □ (sb) 최종적으로 융합 특징과 예측된 3D 키포인트를 concat하여 드론의 rotation과 translation을 독립적인 헤드를 통해 추정한다.
        %Finally, we concatenate the fused features with the predicted 3D keypoints and estimate the drone's rotation and translation through separate pose heads.
        Finally, we concatenate the fused features with the predicted 3D keypoints and feed them into an MLP-based pose head as follows:
        \begin{equation}
            \mathbf{y}^{pose} = \text{MLP}_{pose}\left(\left[\mathbf{f}_{fused}, \ \mathbf{y}^{3D}\right]\right).
        \end{equation}
    % ⭐ (sb) 회전은 3개의 각도를 추정하고, 각 값을 sigmoid 활성화를 통해 [0,1] 범위로 정규화하여 네트워크 학습에서 더 안정적이고 효율적인 수렴을 달성한다 \cite{lecun1998efficient}.  
        %For rotation, we estimate three angles and normalize each value to the range $[0,1]$ through sigmoid activation, which provides more stable and efficient convergence during training\textcolor{skyblue}{\cite{lecun1998efficient}}.
    % ㅁ (sb) Translation은 카메라 좌표계에서의 3차원 벡터를 직접 회귀한다.
        %Translation is directly regressed as a 3D vector in the camera coordinate system.\
        The details of the MLPs ($\text{MLP}_{3D}$ and $\text{MLP}_{pose}$) are illustrated in Fig.~\textcolor{skyblue}{\ref{fig:02}}b.
        The output 6-dimensional vector $\mathbf{y}^{pose} \in \mathbb{R}^6$ is split into rotation and translation components. 
        The first three elements are processed through sigmoid activation to obtain normalized rotation angles $\mathbf{r}^{pred}$, which provides more stable and efficient convergence during training~\cite{lecun1998efficient}.
        The last three elements directly represent the translation vector $\mathbf{t}^{pred}$ in the camera coordinate system.

    \subsection{Loss Functions}
    \label{sec:methods:loss}
        % ■ (sb) Loss 개괄
            % □ (sb) 우리 방법은 end-to-end 학습으로, Keypoint Encoder와 Pose Decoder를 함께 최적화한다.
            % □ (sb) 두 모듈은 서로 연관된 과제를 수행하며, 손실 함수는 이를 함께 반영한다. 
            % □ (sb) 이를 통해 단계 간 일관된 학습이 가능하다.
                % Our framework is trained end-to-end, so the keypoint encoder and 3D pose decoder are optimized together.
                % Both modules solve related tasks, and the loss combines their objectives. 
                % This joint design ensures consistent learning across stages.
               Our framework is trained end-to-end, where the keypoint encoder and 3D pose decoder are optimized jointly. Since both modules address closely related tasks, we design a unified loss that combines their objectives, ensuring consistent learning across stages.
       % ■ (sb) encoder loss
           % □ (sb) 그래서 Keypoint Encoder는 2D 키포인트 검출과 드론 클래스 분류를 동시에 수행하는 멀티태스크 학습 구조로 설계되었다.
               The keypoint encoder is formulated as a multi-task structure that performs 2D keypoint detection and drone class classification simultaneously.
               % To this end, we first describe the keypoint encoder, which is formulated as a multi-task learning structure that performs 2D keypoint detection and drone class classification simultaneously.
           % □ (sb) 2D 키포인트 검출을 위해서는 예측된 키포인트 좌표와 ground truth 간의 Mean Squared Error (MSE) 손실을 사용한다.
            % -> DroneKey는 Pnp라 2D keypoint가 잘 검출되어야 해서 pose adaptive한 loss를 사용했지만, end-to-end 구조인 DroneKey++에서는 3D pose 추정이 최종 목표로, 단순한 MSE loss를 사용했습니다...!
               For 2D keypoint detection, we use mean squared error (MSE) loss $\mathcal{L}_{\text{2D}}$ between predicted keypoint coordinates $\mathbf{y}^{2D}$ and ground truth.
           % □ (sb) 드론 클래스 분류를 위해서는 예측된 확률 분포와 실제 클래스 레이블 간의 cross-entropy 손실을 사용한다.
               For drone class classification, we use cross-entropy loss $\mathcal{L}_{\texttt{cls}}$ between predicted probability distribution $\mathbf{y}^\texttt{cls}$ and actual class labels.
           % □ (sb) Keypoint Encoder 손실 함수는 다음과 같이 정의된다:
               The loss function for the keypoint encoder is defined as follows:
               \begin{equation}
                   \mathcal{L}_{\text{enc}} = \mathcal{L}_{\text{2D}} + \mathcal{L}_{\texttt{cls}}.
               \end{equation}
           % % □ (sb) 여기서 $\mathcal{L}_{3D}$는 3D 키포인트 예측을 위한 MSE 손실이다.
           %     where $\mathcal{L}_{3D}$ is MSE loss for 3D keypoint prediction.
           
       % ■ (sb) pose decoder loss
           % □ (sb) Pose Decoder는 3D 키포인트 예측, 회전과 평행이동을 예측하는 회귀 태스크를 수행한다.
               The 3D pose decoder is trained through regression tasks to predict 3D keypoints, rotation, and translation, and is optimized with corresponding loss functions.
               The loss for 3D keypoint prediction is denoted as $\mathcal{L}_{\text{3D}}$, while the loss for translation is denoted as $\mathcal{L}_{\text{trans}}$, both computed as mean squared error (MSE) between predictions and ground truth.
               Here, the 3D keypoints correspond to the coordinates of the four drone propellers, whereas the translation represents the 3D coordinates of the drone’s center.
           % □ (sb)  3D loss는 3D 키포인트 예측을 위한 MSE 손실이다.
               % $\mathcal{L}_{\text{3D}}$ is defined as the mean squared error (MSE) loss between $\mathbf{y}^{3D}$ and ground-truth position for 3D keypoint prediction.
           % □ (sb) 회전의 경우 정규화된 회전 표현의 주기적 특성을 고려한 순환 손실 함수를 적용한다.
             % □ (sb) 예측된 회전 $\mathbf{R}$와 실제 회전 $\mathbf{R}^{gt}$가 모두 $[0, 1]$ 범위로 정규화되었을 때, 순환 손실 함수는 다음과 같이 정의된다:
               % For rotation, we apply a rotation loss function that considers the periodic characteristics of normalized rotation representation.
               % When both predicted rotation $\mathbf{r}^{pred}$ and ground truth rotation $\mathbf{r}^{gt}$ are normalized to the $[0, 1]$ range, the circular loss function is defined as follows:
               For rotation, we design a loss function that accounts for the periodic nature of rotation representations.
               Both the predicted rotation $\mathbf{r}^{pred}$ and the ground-truth rotation $\mathbf{r}^{gt}$ are normalized to the $[0,1]$ range, and the circular loss is then defined as:
               \begin{equation}
               \small
                   \mathcal{L}_{\text{rot}} = \frac{1}{3} \sum_{j \in \{x,y,z\}} \left[\min\left(|\mathbf{r}^{pred}_j - \mathbf{r}^{gt}_j|, 1 - |\mathbf{r}^{pred}_j - \mathbf{r}^{gt}_j|\right)\right]^2.
               \end{equation}
           % □ (sb) 이 공식은 단위원 상에서 가장 짧은 각도 거리를 직접 계산하며, 0과 1이 동일한 각도 위치를 의미함을 반영한다.
               This formula directly computes the shortest angular distance on the unit circle and reflects that 0 and 1 represent the same angular position.
            % □ (sb) 평행이동 손실의 경우 일반적인 MSE 손실을 사용한다.
           % □ (sb) 최종적으로 Pose Decoder 손실 함수는 회전 손실과 평행이동 손실을 결합하여 다음과 같이 정의된다:
              The 3D pose decoder loss function is defined by combining the losses as follows:
               \begin{equation}
                   \mathcal{L}_{\text{dec}} = \mathcal{L}_{\text{3D}} + \mathcal{L}_{\text{rot}} + \mathcal{L}_{\text{trans}}.
               \end{equation}

        % ■ (sb) total loss
            % □ (sb) 마지막으로 Keypoint Encoder와 Pose Decoder에서 정의된 손실들을 단순 합산하여 최종 학습 손실을 구성한다.
            % □ (sb) 즉, 최종 손실 함수는 다음과 같이 정의된다:
                Finally, the overall training objective is obtained by summing the encoder and decoder losses:
                \begin{equation}
                    \mathcal{L}_{\text{total}} = \mathcal{L}_{\text{enc}} + \mathcal{L}_{\text{dec}}.
                \end{equation}
            % □ (sb) 본 연구에서는 다양한 가중치 조합을 실험적으로 검증하였으며, 최종적으로는 단순 합산 방식이 가장 안정적이고 효과적인 성능을 보임을 확인하였다.
                In this work, we experimentally investigated several weighting strategies for different loss terms and confirmed that the simple summation scheme consistently yielded stable and effective performance. 
            % □ (sb) 이에 대한 세부적인 비교와 검증 결과는 Section~\ref{sec:experiments}에 제시하였다.
                A detailed comparison and validation of these weighting strategies is provided in Sec.~\textcolor{skyblue}{\ref{sec:exp:loss}}.

%■■■■■■■■■■■■■■■■■■■■■■■■■■■■■■■■■■■■■■■■■■■■■■■■■■■■■■■■■■■■■■■■■■■■■■■■■■■
\section{Dataset - \texttt{6DroneSyn}}
\label{sec:dataset}

    % ✅ (yjcho) subsection 삭제
    %\subsection{Overview and Statistics}
    %\label{sec:dataset:overview}
        % □ (sb) 본 연구에서는 드론의 3D 자세 추정을 위한 종합적인 벤치마크 데이터셋인 \texttt{6DroneSyn}을 구축하였다.
            We present \texttt{6DroneSyn}, a comprehensive benchmark dataset for drone 3D pose estimation.
        % □ (sb) 데이터셋은 총 52,920장의 고해상도(1920×1080) 합성 이미지를 포함한다.
            The dataset contains 52,920 high-resolution (1920×1080) synthetic images.
        % □ (sb) 전체 데이터셋은 120GB+의 대용량으로 구성되어 기존 드론 자세 추정 데이터셋들 대비 상당한 규모를 가진다.
            The entire dataset comprises 120GB+, representing a substantial scale compared to existing drone pose estimation datasets.
        % □ (sb) 데이터셋은 7종의 DJI 드론 모델(Mini3 Pro, Mini2, Air3, Air2, Mavic 2 Pro, Mavic3, Tello)과 88개의 다양한 배경 환경을 사용하여 실제 감시 환경의 다양성을 반영한다.
            It contains 7 DJI drone models (Mini3 Pro, Mini2, Air3, Air2, Mavic 2 Pro, Mavic3, Tello) and 88 diverse background environments to reflect the variety of real surveillance scenarios.
        % □ (sb) 합성 데이터는 13개의 scene로 구성되며, 각 scene마다 드론별로 독립적인 sequence를 생성하여 총 91개의 sequence를 포함한다. 
            The synthetic data comprises 13 scenes, with each scene generating independent sequences for each drone model, totaling 91 sequences. 
        % □ (sb) 이들은 단순한 선형 움직임부터 복잡한 3차원 궤적까지 다양한 비행 패턴을 포괄하여 실제 운용 상황에서의 모든 유형의 드론 움직임을 커버한다.
            These cover various flight patterns, from simple linear movements to complex 3D trajectories, encompassing all types of drone motions in real operations.
            The summary and overall composition of \texttt{6DroneSyn} are presented in Tab.~\textcolor{skyblue}{\ref{tab:02}}.

\begin{figure}[t]
    % \vspace{-5pt}
	\centering
        \subfigure[2D annotations and an example frame from the sequence]{\includegraphics[width=1\columnwidth]{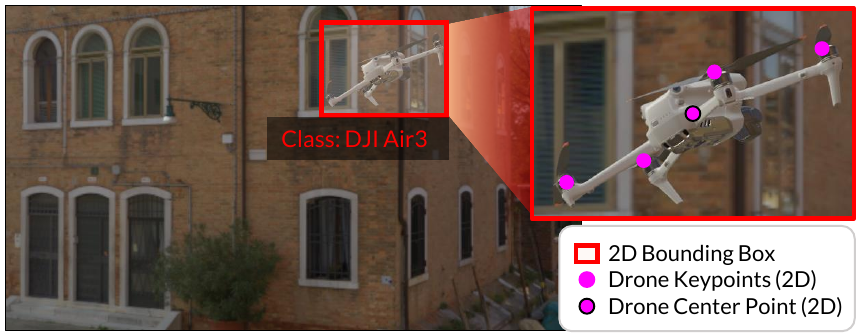}}
		\subfigure[3D annotations]{\includegraphics[width=0.4655\columnwidth]{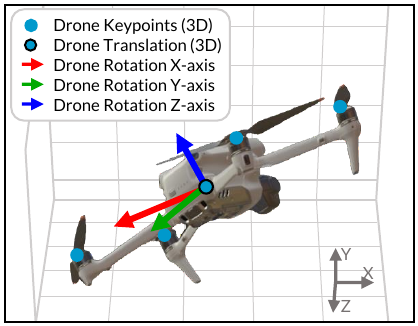}}\hspace{10pt}
		\subfigure[3D trajectory]{\includegraphics[width=0.4655\columnwidth]{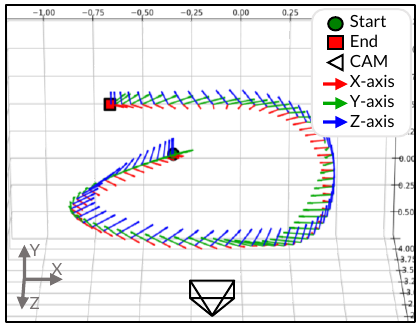}} \vspace{-5pt}
    	\caption{\textbf{\texttt{6DroneSyn} Dataset Annotations.} 
                \textls[-20]{
                Examples of annotations: (a) 2D bounding boxes and 2D keypoints, 
                (b) 3D keypoints with translation and rotation vectors, 
                and (c) full 3D trajectories with rotation and translation from sequential images.}}
	\label{fig:03}  \vspace{-10pt}
    \end{figure}

    % ✅ (yjcho) subsection 삭제
    % \subsection{Data Generation and Annotation}
    % \label{sec:dataset:annotation}
        % \subsubsection{Synthetic Data Generation (\texttt{6DroneSyn})}
        % □ (sb) 합성 데이터 생성을 위해 22개의 실제 환경에서 수집된 360-degree 배경 이미지를 활용하고, 각 환경을 Z축 기준으로 90도씩 회전하여 총 88개의 배경 변형을 확보하였다.
            For synthetic data generation, we utilized 360-degree background images collected from 22 real environments. 
            Each environment was rotated 90 degrees around the Z-axis to create a total of 88 background variations. 
            Several examples of the backgrounds are shown in Fig.~\textcolor{skyblue}{\ref{fig:01}}c.
        % □ (sb) 7종의 DJI 드론 3D 모델을 사용하여 소형부터 전문가용까지의 형태적 다양성을 보장하였다.
            We used 7 DJI 3D drone models covering a wide range from small to professional types.
        % □ (sb) 각 프레임에는 다음과 같은 정밀한 어노테이션 정보가 포함된다:
            Each frame contains the following precise annotations, as illustrated in Fig.~\textcolor{skyblue}{\ref{fig:03}}:
        \begin{itemize}
            % \item 드론 프로펠러의 2D 키포인트 좌표 및 대응하는 3D 월드 좌표와 카메라 좌표,
            \item 2D keypoint coordinates of drone propellers and corresponding 3D coordinates.
            % \item 완전한 3D 자세 정보 (3D translation + 3D rotation),
            \item Drone 3D pose (rotation $\mathbf{R}$ and translation $\mathbf{t}$).
            % \item 카메라 내부/외부 파라미터,
            \item Camera intrinsic and extrinsic parameters.
            % \item 드론 영역의 바운딩 박스.
            \item 2D bounding boxes of drone regions in full-frame videos.
        \end{itemize}
        % □ (sb) 모든 어노테이션은 렌더링 파이프라인에서 자동 생성되어 인간의 라벨링 오류를 완전히 배제하였다.
            All annotations are automatically generated in the rendering pipeline, completely eliminating human labeling errors.
            More details of our dataset are provided in the supplementary video.

\begin{table}[t]
    \setlength{\tabcolsep}{4pt} % 기본은 6pt
    \centering
    \caption{
    \textbf{\texttt{6DroneSyn} dataset composition.} 
        The dataset consists of 13 synthetic scenes, each featuring diverse drone motion patterns (linear and non-linear) and 22 background variations. Each scene is further divided into 21 subsequences, created from 7 drone types and 3 backgrounds.}
    \label{tab:02}
        \begin{tabular}{cccccc}
        \hline \noalign{\hrule height 0.5pt}
        \textbf{Scene}  & \textbf{Motion}  & \textbf{\# of Subseq.}  & \textbf{FPS} & \textbf{Duration} & \textbf{Total frames} \\ \hline \noalign{\hrule height 0.5pt}
        \#01--\#03      & Linear           & 21                      & 30              & 4s             & 7,560     \\
        \#04--\#06      & Non-linear       & 21                      & 30              & 4s             & 7,560     \\
        \#07--\#09      & Linear           & 21                      & 30              & 4s             & 7,560     \\
        \#10--\#13      & Non-linear       & 21                      & 30              & 12s            & 30,240    \\ \hline  \noalign{\hrule height 0.5pt}
        %\multicolumn{8}{c}{All}                                                                                                                                                    & \textbf{52,920} \\ \hline \noalign{\hrule height 0.5pt}
        \end{tabular}
    \end{table}

%■■■■■■■■■■■■■■■■■■■■■■■■■■■■■■■■■■■■■■■■■■■■■■■■■■■■■■■■■■■■■■■■■■■■■■■■■■■
    
    \begin{table*}[t!]
    \setlength{\tabcolsep}{2pt} % 기본은 6pt
    \centering
    \caption{
    \textls[-10]{\textbf{Performance comparison with drone 3D pose estimation approaches.} 
    Results on scene \#06 and scene \#07 show that our method achieves the best accuracy in both rotation and translation estimation, outperforming DronePose~\cite{DronePose} and DroneKey~\cite{DroneKey}.}}

    \label{tab:03}  

\begin{tabular}{cc|cc|cccccccccccccc|c}
\hline \noalign{\hrule height 0.5pt}
\multicolumn{2}{c|}{}                                                                                                                                                                                                     & \multicolumn{2}{c|}{\textbf{Metric}}                                               & \multicolumn{14}{c|}{\textbf{Test scenes}}                                                                                                                                                                                                                                                                                                                                                                                                                                                                                                                                                                                                                                                                                                                                                                                                                                 &                                                               \\ \cline{3-18}
\multicolumn{2}{c|}{}                                                                                                                                                                                                     &                                                &                                   & \multicolumn{7}{c|}{\textbf{Scene \#06}}                                                                                                                                                                                                                                                                                                                                                                                                                                           & \multicolumn{7}{c|}{\textbf{Scene \#07}}                                                                                                                                                                                                                                                                                                                                              &                                                               \\
\multicolumn{2}{c|}{\multirow{-3}{*}{\textbf{Methods}}}                                                                                                                                                                   & \multirow{-2}{*}{\textbf{3D Pose}}             & \multirow{-2}{*}{\textbf{Metric}} & \textbf{Mini3}                                                & \textbf{Mini2}                                                & \textbf{Air3}                                                 & \textbf{Air2}                                                 & \textbf{Mav2}                                                 & \textbf{Mav3}                                                 & \multicolumn{1}{c|}{\textbf{Tello}}                                                & \textbf{Mini3}                                                & \textbf{Mini2}                                                & \textbf{Air3}                         & \textbf{Air2}                         & \textbf{Mav2}                         & \textbf{Mav3}                                                 & \textbf{Tello}                                                & \multirow{-3}{*}{\textbf{Average}}                            \\ \hline \noalign{\hrule height 0.5pt}
\multicolumn{1}{c|}{}                                                                                    &                                                                                                                &                                                & MAE                               & 90.55                                                         & {\color[HTML]{333333} 69.89}                                  & {\color[HTML]{333333} 90.28}                                  & 90.45                                                         & 78.51                                                         & 91.39                                                         & \multicolumn{1}{c|}{69.52}                                                         & 99.79                                                         & {\color[HTML]{333333} 64.31}                                  & {\color[HTML]{333333} 60.17}          & {\color[HTML]{333333} 74.37}          & {\color[HTML]{333333} 90.85}          & 74.44                                                         & 76.92                                                         & 79.25                                                         \\
\multicolumn{1}{c|}{}                                                                                    &                                                                                                                & \multirow{-2}{*}{$\mathbf{R}$ $(^\circ)$}                         & MedAE                             & 98.98                                                         & 84.75                                                         & {\color[HTML]{333333} 90.26}                                  & 96.51                                                         & 76.33                                                         & 108.96                                                        & \multicolumn{1}{c|}{65.03}                                                         & 106.11                                                        & {\color[HTML]{333333} 78.11}                                  & {\color[HTML]{333333} 87.03}          & {\color[HTML]{333333} 86.94}          & {\color[HTML]{333333} 96.92}          & 62.46                                                         & 70.06                                                         & 87.55                                                         \\
\multicolumn{1}{c|}{}                                                                                    &                                                                                                                &                                                & MAE                               & 0.656                                                         & 0.295                                                         & 0.121                                                         & 0.343                                                         & 0.450                                                         & 0.426                                                         & \multicolumn{1}{c|}{{\color[HTML]{333333} 0.203}}                                  & 0.521                                                         & 0.313                                                         & 0.184                                 & {\color[HTML]{333333} 0.448}          & {\color[HTML]{333333} 0.670}          & 0.710                                                         & 0.199                                                         & 0.391                                                         \\
\multicolumn{1}{c|}{}                                                                                    & \multirow{-4}{*}{\begin{tabular}[c]{@{}c@{}}Keypoint\\ Detector~\cite{yolov8}\\ +\\ PnP~\cite{pnp}\end{tabular}}                        & \multirow{-2}{*}{$\mathbf{t}$ (m)}                          & MedAE                             & 0.280                                                         & \textbf{0.088}                                                & \textbf{0.027}                                                & 0.095                                                         & \textbf{0.123}                                                & {\color[HTML]{333333} \textbf{0.131}}                         & \multicolumn{1}{c|}{{\color[HTML]{333333} 0.060}}                                  & \textbf{0.155}                                                & \textbf{0.094}                                                        & 0.031                                 & {\color[HTML]{333333} 0.120}          & {\color[HTML]{333333} 0.154}          & {\color[HTML]{333333} \textbf{0.128}}                         & 0.056                                                         & \textbf{0.105}                                                \\ \cline{2-19} 
\multicolumn{1}{c|}{}                                                                                    &                                                                                                                &                                                & MAE                               & 33.84                                                         & {\color[HTML]{333333} \textbf{14.61}}                         & {\color[HTML]{333333} \textbf{11.51}}                         & 24.78                                                         & 22.25                                                         & 31.85                                                         & \multicolumn{1}{c|}{32.13}                                                         & 25.32                                                         & {\color[HTML]{333333} \textbf{4.660}}                         & {\color[HTML]{333333} \textbf{8.290}} & {\color[HTML]{333333} \textbf{7.350}} & {\color[HTML]{333333} \textbf{21.95}} & 22.29                                                         & 21.49                                                         & 20.17                                                         \\
\multicolumn{1}{c|}{}                                                                                    &                                                                                                                & \multirow{-2}{*}{$\mathbf{R}$ $(^\circ)$}                         & MedAE                             & 31.96                                                         & {\color[HTML]{333333} 14.06}                                  & {\color[HTML]{333333} \textbf{1.840}}                         & 24.35                                                         & 20.61                                                         & 33.48                                                         & \multicolumn{1}{c|}{30.70}                                                         & 26.98                                                         & {\color[HTML]{333333} \textbf{1.580}}                         & {\color[HTML]{333333} \textbf{8.540}} & {\color[HTML]{333333} \textbf{6.400}} & {\color[HTML]{333333} \textbf{25.15}} & 23.20                                                         & 21.17                                                         & 21.28                                                         \\
\multicolumn{1}{c|}{}                                                                                    &                                                                                                                &                                                & MAE                               & 0.668                                                         & 1.094                                                         & 1.005                                                         & 0.211                                                         & 0.462                                                         & 0.208                                                         & \multicolumn{1}{c|}{{\color[HTML]{333333} \textbf{0.056}}}                         & {\color[HTML]{333333} 0.484}                                  & 0.952                                                         & {\color[HTML]{333333} 0.911}          & {\color[HTML]{333333} \textbf{0.017}} & {\color[HTML]{333333} \textbf{0.168}} & {\color[HTML]{333333} 0.648}                                  & 0.577                                                         & 0.533                                                         \\
\multicolumn{1}{c|}{\multirow{-8}{*}{\textbf{\begin{tabular}[c]{@{}c@{}}Multi-\\ stage\end{tabular}}}}   & \multirow{-4}{*}{DroneKey~\cite{DroneKey}}                                                                                     & \multirow{-2}{*}{$\mathbf{t}$ (m)}                          & MedAE                             & 0.656                                                         & 0.830                                                         & 1.002                                                         & \textbf{0.032}                                                & 0.235                                                         & {\color[HTML]{333333} 0.187}                                  & \multicolumn{1}{c|}{{\color[HTML]{333333} \textbf{0.056}}}                         & {\color[HTML]{333333} 0.478}                                  & 0.787                                                         & {\color[HTML]{333333} 0.950}          & {\color[HTML]{333333} \textbf{0.017}} & {\color[HTML]{333333} \textbf{0.112}} & {\color[HTML]{333333} 0.221}                                  & 0.048                                                         & 0.370                                                         \\ \hline \noalign{\hrule height 0.5pt}
\multicolumn{1}{c|}{}                                                                                    &                                                                                                                &                                                & MAE                               & 80.33                                                         & {\color[HTML]{333333} 23.67}                                  & {\color[HTML]{333333} 135.5}                                  & 45.49                                                         & 86.16                                                         & 113.9                                                         & \multicolumn{1}{c|}{88.17}                                                         & 79.42                                                         & {\color[HTML]{333333} 15.27}                                  & {\color[HTML]{333333} 121.5}          & {\color[HTML]{333333} 85.72}          & {\color[HTML]{333333} 82.72}          & 105.0                                                         & 49.45                                                         & 79.06                                                         \\
\multicolumn{1}{c|}{}                                                                                    &                                                                                                                & \multirow{-2}{*}{$\mathbf{R}$ $(^\circ)$}                         & MedAE                             & 81.56                                                         & {\color[HTML]{333333} \textbf{8.400}}                         & {\color[HTML]{333333} 148.4}                                  & 32.47                                                         & 100.1                                                         & 135.5                                                         & \multicolumn{1}{c|}{103.3}                                                         & 100.4                                                         & {\color[HTML]{333333} 9.650}                                  & {\color[HTML]{333333} 126.0}          & {\color[HTML]{333333} 88.95}          & {\color[HTML]{333333} 69.28}          & 109.3                                                         & 43.17                                                         & 82.63                                                         \\
\multicolumn{1}{c|}{}                                                                                    &                                                                                                                &                                                & MAE                               & 0.148                                                         & 0.161                                                         & 0.108                                                         & 0.122                                                         & 0.176                                                         & 0.180                                                         & \multicolumn{1}{c|}{{\color[HTML]{333333} 0.163}}                                  & {\color[HTML]{333333} \textbf{0.221}}                         & 0.179                                                         & {\color[HTML]{333333} \textbf{0.132}} & {\color[HTML]{333333} \textbf{0.134}} & {\color[HTML]{333333} \textbf{0.173}} & {\color[HTML]{333333} \textbf{0.153}}                         & 0.236                                                         & 0.163                                                         \\ 
\multicolumn{1}{c|}{}                                                                                    & \multirow{-4}{*}{DronePose~\cite{DronePose}}                                                                                    & \multirow{-2}{*}{$\mathbf{t}$ (m)}                          & MedAE                             & 0.280                                                         & 0.322                                                         & 0.169                                                         & 0.219                                                         & 0.289                                                         & {\color[HTML]{333333} 0.353}                                  & \multicolumn{1}{c|}{{\color[HTML]{333333} 0.301}}                                  & {\color[HTML]{333333} 0.497}                                  & 0.385                                                         & {\color[HTML]{333333} \textbf{0.254}} & {\color[HTML]{333333} \textbf{0.246}} & {\color[HTML]{333333} \textbf{0.328}} & {\color[HTML]{333333} 0.238}                                  & 0.536                                                         & 0.317                                                         \\ \cline{2-19} 
\multicolumn{1}{c|}{}                                                                                    & \cellcolor[HTML]{EFEFEF}                                                                                       & \cellcolor[HTML]{EFEFEF}                       & \cellcolor[HTML]{EFEFEF}MAE       & \cellcolor[HTML]{EFEFEF}{\color[HTML]{333333} \textbf{9.910}} & \cellcolor[HTML]{EFEFEF}19.91                                 & \cellcolor[HTML]{EFEFEF}19.74                                 & \cellcolor[HTML]{EFEFEF}{\color[HTML]{333333} \textbf{18.27}} & \cellcolor[HTML]{EFEFEF}{\color[HTML]{333333} \textbf{19.36}} & \cellcolor[HTML]{EFEFEF}{\color[HTML]{333333} \textbf{12.09}} & \multicolumn{1}{c|}{\cellcolor[HTML]{EFEFEF}{\color[HTML]{333333} \textbf{14.42}}} & \cellcolor[HTML]{EFEFEF}{\color[HTML]{333333} \textbf{16.61}} & \cellcolor[HTML]{EFEFEF}22.35                                 & \cellcolor[HTML]{EFEFEF}15.53         & \cellcolor[HTML]{EFEFEF}17.08         & \cellcolor[HTML]{EFEFEF}26.90         & \cellcolor[HTML]{EFEFEF}{\color[HTML]{333333} \textbf{12.49}} & \cellcolor[HTML]{EFEFEF}{\color[HTML]{333333} \textbf{18.32}} & \cellcolor[HTML]{EFEFEF}{\color[HTML]{333333} \textbf{17.34}} \\
\multicolumn{1}{c|}{}                                                                                    & \cellcolor[HTML]{EFEFEF}                                                                                       & \multirow{-2}{*}{\cellcolor[HTML]{EFEFEF}$\mathbf{R}$ $(^\circ)$} & \cellcolor[HTML]{EFEFEF}MedAE     & \cellcolor[HTML]{EFEFEF}{\color[HTML]{333333} \textbf{8.580}} & \cellcolor[HTML]{EFEFEF}20.38                                 & \cellcolor[HTML]{EFEFEF}18.68                                 & \cellcolor[HTML]{EFEFEF}{\color[HTML]{333333} \textbf{16.82}} & \cellcolor[HTML]{EFEFEF}{\color[HTML]{333333} \textbf{17.20}} & \cellcolor[HTML]{EFEFEF}{\color[HTML]{333333} \textbf{11.76}} & \multicolumn{1}{c|}{\cellcolor[HTML]{EFEFEF}{\color[HTML]{333333} \textbf{14.01}}} & \cellcolor[HTML]{EFEFEF}{\color[HTML]{333333} \textbf{16.85}} & \cellcolor[HTML]{EFEFEF}23.24                                 & \cellcolor[HTML]{EFEFEF}15.13         & \cellcolor[HTML]{EFEFEF}16.44         & \cellcolor[HTML]{EFEFEF}27.27         & \cellcolor[HTML]{EFEFEF}{\color[HTML]{333333} \textbf{12.38}} & \cellcolor[HTML]{EFEFEF}{\color[HTML]{333333} \textbf{19.73}} & \cellcolor[HTML]{EFEFEF}{\color[HTML]{333333} \textbf{17.10}} \\
\multicolumn{1}{c|}{}                                                                                    & \cellcolor[HTML]{EFEFEF}                                                                                       & \cellcolor[HTML]{EFEFEF}                       & \cellcolor[HTML]{EFEFEF}MAE       & \cellcolor[HTML]{EFEFEF}{\color[HTML]{333333} \textbf{0.121}} & \cellcolor[HTML]{EFEFEF}{\color[HTML]{333333} \textbf{0.142}} & \cellcolor[HTML]{EFEFEF}{\color[HTML]{333333} \textbf{0.067}} & \cellcolor[HTML]{EFEFEF}{\color[HTML]{333333} \textbf{0.100}} & \cellcolor[HTML]{EFEFEF}{\color[HTML]{333333} \textbf{0.140}} & \cellcolor[HTML]{EFEFEF}{\color[HTML]{333333} \textbf{0.114}} & \multicolumn{1}{c|}{\cellcolor[HTML]{EFEFEF}0.075}                                 & \cellcolor[HTML]{EFEFEF}0.268                                 & \cellcolor[HTML]{EFEFEF}{\color[HTML]{333333} \textbf{0.149}} & \cellcolor[HTML]{EFEFEF}0.135         & \cellcolor[HTML]{EFEFEF}0.080         & \cellcolor[HTML]{EFEFEF}0.234         & \cellcolor[HTML]{EFEFEF}0.187                                 & \cellcolor[HTML]{EFEFEF}{\color[HTML]{333333} \textbf{0.083}} & \cellcolor[HTML]{EFEFEF}{\color[HTML]{333333} \textbf{0.135}} \\
\multicolumn{1}{c|}{\multirow{-8}{*}{\textbf{\begin{tabular}[c]{@{}c@{}}End-\\ to-\\ end\end{tabular}}}} & \multirow{-4}{*}{\cellcolor[HTML]{EFEFEF}\textbf{\begin{tabular}[c]{@{}c@{}}DroneKey\textcolor{encoderpink}{+}\textcolor{decoderblue}{+}\\ (Ours)\end{tabular}}} & \multirow{-2}{*}{\cellcolor[HTML]{EFEFEF}$\mathbf{t}$ (m)}  & \cellcolor[HTML]{EFEFEF}MedAE     & \cellcolor[HTML]{EFEFEF}{\color[HTML]{333333} \textbf{0.224}} & \cellcolor[HTML]{EFEFEF}{\color[HTML]{333333} 0.175}          & \cellcolor[HTML]{EFEFEF}{\color[HTML]{333333} 0.129}          & \cellcolor[HTML]{EFEFEF}{\color[HTML]{333333} 0.181}          & \cellcolor[HTML]{EFEFEF}{\color[HTML]{333333} 0.248}          & \cellcolor[HTML]{EFEFEF}0.219                                 & \multicolumn{1}{c|}{\cellcolor[HTML]{EFEFEF}0.158}                                 & \cellcolor[HTML]{EFEFEF}0.561                                 & \cellcolor[HTML]{EFEFEF}{\color[HTML]{333333} 0.187} & \cellcolor[HTML]{EFEFEF}0.293         & \cellcolor[HTML]{EFEFEF}0.150         & \cellcolor[HTML]{EFEFEF}0.580         & \cellcolor[HTML]{EFEFEF}0.391                                 & \cellcolor[HTML]{EFEFEF}{\color[HTML]{333333} \textbf{0.171}} & \cellcolor[HTML]{EFEFEF}{\color[HTML]{333333} 0.242}          \\ \hline \noalign{\hrule height 0.5pt}
\end{tabular}
    
    %\\[0.15cm]
    %    † retrained without 3D mesh priors. \vspace{-10pt}
    \vspace{-10pt}
    \end{table*}

%■■■■■■■■■■■■■■■■■■■■■■■■■■■■■■■■■■■■■■■■■■■■■■■■■■■■■■■■■■■■■■■■■■■■■■■■■■■

\section{EXPERIMENTAL RESULTS}
\label{sec:experimental_results}
% 실험은 과거형으로 써야댐
    
    \subsection{Experimental Settings}
        %\subsubsection{\textbf{Datasets}}
        \noindent \textbf{Datasets.} \quad
            % ㅁ (sb) 모든 실험은 제안한 \texttt{6Drone} 데이터셋을 사용하였다. 
                All experiments were conducted on the proposed \texttt{6DroneSyn} dataset. 
            % ㅁ (sb) 데이터 크기는 약 120GB로, 기존 드론 자세 추정 데이터셋보다 상당히 크다. 
                The dataset size is about 120GB, significantly larger than existing drone pose datasets. 
            % ㅁ (sb) 데이터셋은 시나리오 단위로 분할되었으며, 학습 (01-02, 04-05, 08, 11-13), 검증 (03, 09, 14), 테스트 (06-07) 세트로 구성되었다. 
                It was split scenario-wise into train (01-02, 04-05, 08, 10-12), validation (03, 09, 13), and test (06-07) sets.
                The validation set was used to select the best model checkpoints during training, and the selected models were subsequently evaluated on the test set. This experimental setup follows the common practice adopted in most related works.
            % % ㅁ (sb) 실제 데이터 (10)는 정성적 평가에 활용되었다. 
            %     with real data (10) used for qualitative evaluation. 

        %\subsubsection{\textbf{Evaluation metrics}}
        \noindent \textbf{Evaluation metrics.} \quad
        % \textcolor{red}{다시 확인하기! -> 논리}
            % ■ (sb) 3D pose 추정은 회전과 위치 정확도를 모두 평가해야 한다.  
            % ■ (sb) 일반적으로 MAE, MedAE, RMSE와 같은 지표가 사용된다.  
            % ■ (sb) 그러나 RMSE는 일부 큰 오차에 민감하여 안정성을 충분히 반영하지 못한다.  
            % ■ (sb) 따라서 본 연구는 평균적인 정확도와 더불어 안정적인 전형적 성능을 보여주기 위해 MAE와 MedAE를 채택한다. 
                %3D pose estimation evaluates both rotation and translation accuracy.  
                %Common metrics include MAE, MedAE, and RMSE~\cite{hodson2022rmse}\cite{stamm2020pose}.  
                %However, RMSE is sensitive to a few large errors and does not capture stability well~\cite{lee2022whats}\cite{lee2024alignment}.  
                3D pose estimation evaluates both rotation and translation accuracy. To assess these aspects, common metrics such as MAE, MedAE, and RMSE are employed~\cite{hodson2022rmse}\cite{stamm2020pose}. 
                However, RMSE is often sensitive to outliers and instability~\cite{lee2022whats}\cite{lee2024alignment}; therefore, we adopt MAE and MedAE to better capture both overall accuracy and stable, representative performance.
            % ■ (sb) 회전 오차는 예측 회전 행렬과 실제 회전 행렬의 각도 차이를 정규화하여 정의한다.  
                %-> IROS에서는 언급하지 않고 MAE 사용했다 정도로만 설명했었습니다!
                The rotation error is defined as the normalized angular difference between the predicted and ground-truth rotation matrices as follows:
                % \begin{align}
                %     \text{MAE}_{\mathbf{r}} &= \frac{1}{S} \sum_{s=1}^{S} 
                %     \frac{1}{\pi}\arccos\!\left(\frac{\mathrm{tr}\!\left((\mathbf{r}_s^{pred})^{\mathsf{T}} \mathbf{r}_s^{gt}\right) - 1}{2}\right), \\
                %     \text{MedAE}_{\mathbf{r}} &= \operatorname{median}_{s}
                %     \left\{
                %     \frac{1}{\pi}\arccos\!\left(\frac{\mathrm{tr}\!\left((\mathbf{r}_s^{pred})^{\mathsf{T}} \mathbf{r}_s^{gt}\right) - 1}{2}\right)
                %     \right\},
                % \end{align}
                  \begin{equation}
                   \text{MAE}_{\mathbf{r}} = \frac{1}{S} \sum_{s=1}^{S} 
                    \frac{1}{\pi}\arccos\left(\frac{\mathrm{tr}\left((\mathbf{r}_s^{pred})^{\mathsf{T}} \mathbf{r}_s^{gt}\right) - 1}{2}\right), 
                  \end{equation}
                  \begin{equation}
                  \small
                      \text{MedAE}_{\mathbf{r}} = \operatorname{median}_{s}
                    \left\{
                    \frac{1}{\pi}\arccos\left(\frac{\mathrm{tr}\left((\mathbf{r}_s^{pred})^{\mathsf{T}} \mathbf{r}_s^{gt}\right) - 1}{2}\right)
                    \right\},
                  \end{equation}
                where $s$ indexes each of the $S$ samples, and $\mathrm{tr}(\cdot)$ denotes the trace, i.e., the sum of the diagonal elements.
            % ■ (sb) $(\mathbf{R}_s^{pred})^{\mathsf{T}} \mathbf{R}_s^{gt}$ 는 상대 회전을 나타낸다.  
                % → 예측된 회전을 뒤집어 실제 회전에 곱하면, 두 회전이 얼마나 다른지(즉 예측을 실제로 맞추기 위해 추가로 회전해야 하는 양)를 얻을 수 있습니다.  
                $(\mathbf{r}_s^{pred})^{\mathsf{T}} \mathbf{r}_s^{gt}$ represents the relative rotation.  
            % ■ (sb) trace 식으로 얻는 각도는 항상 $0^\circ \sim 180^\circ$ 범위의 최소 회전 차이다.  
                % → 회전 행렬의 성질상 각도 차이는 항상 가장 작은 값으로 표현되며, 축을 뒤집어도 같은 회전이 되기 때문에 최대 180$^\circ$까지만 측정됩니다.
                The trace formula gives the minimal angle difference, always in $0^\circ$–$180^\circ$.  
            % ■ (sb) 이를 $\pi$로 나누어 $[0,1]$로 정규화하며, 1은 $180^\circ$ 차이를 뜻한다.  
                % → 라디안 단위(0~π)를 그대로 쓰면 크기가 크므로, π로 나누어 0~1 사이 값으로 바꾸어 비교가 쉽도록 만든 것입니다.
                Dividing by $\pi$ normalizes the error to the range $[0,1]$, where a value of 1 corresponds to a $180^\circ$ rotation error.
            % ■ (sb) 위치 오차는 예측된 이동 벡터와 실제 이동 벡터 간의 유클리드 거리로 측정한다.  
                Translation error is measured as the Euclidean distance between the predicted and ground-truth translation vectors, evaluated using MAE and MedAE.

        %\subsubsection{\textbf{Implementation details}}
        \noindent \textbf{Implementation details.} \quad
            % ㅁ (sb) 학습 및 평가 환경은 NVIDIA A100 GPU 기반이며, 실시간 성능은 Intel Xeon Gold 6440 CPU에서 추가적으로 확인하였다. 
                Training and evaluation were performed on an NVIDIA A100 GPU, 
                with real-time tests additionally on an Intel Xeon Gold 6440 CPU. 
            % ㅁ (sb) Adam 옵티마이저(초기 학습률 1e-5, cosine annealing 스케줄러)를 사용하였고, 배치 크기는 32, 학습 epoch 수는 100으로 설정하였다. 
                We used the Adam optimizer (initial lr = 1e-5) with a cosine annealing scheduler, 
                batch size 32, and 100 training epochs. 
            % ㅁ (sb) 백본은 ImageNet 사전 학습된 ResNet으로 초기화하였으며, 데이터 증강은 적용하지 않았다. 
                The backbone was initialized with ImageNet~\cite{imagenet} pre-trained ResNet~\cite{resnet}, 
                and no data augmentation was applied. 
            % ㅁ (sb) 모든 비교는 동일한 설정에서 수행되었다. 
                All comparisons were performed under identical settings.

    % □□□□□□□□□□□□□□□□□□□□□□□□□□□□□□□□□□□□□□□□□□□□□□□□□□□□□□□□□□□□□□□□□□□□□□□□□□□□□
    \begin{figure*}[t]
    % \vspace{-5pt}
	\centering
        \subfigure[Scene \#03 (Valid) - Mini3]{\includegraphics[width=0.5\columnwidth]{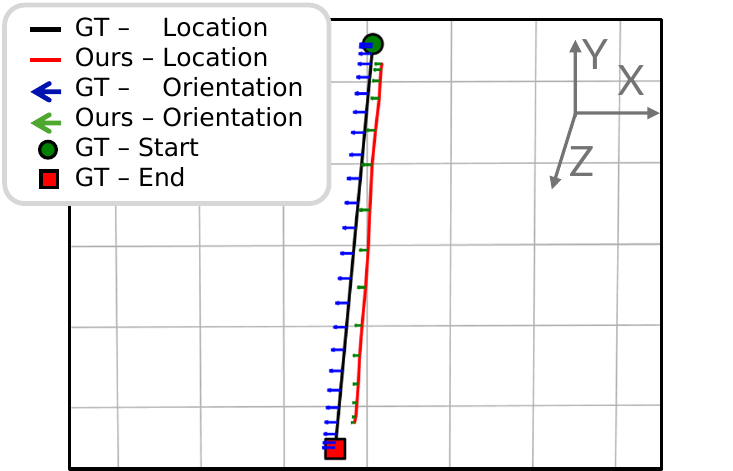}}%\hspace{5pt}
		\subfigure[Scene \#09 (Valid) - Mini2]{\includegraphics[width=0.5\columnwidth]{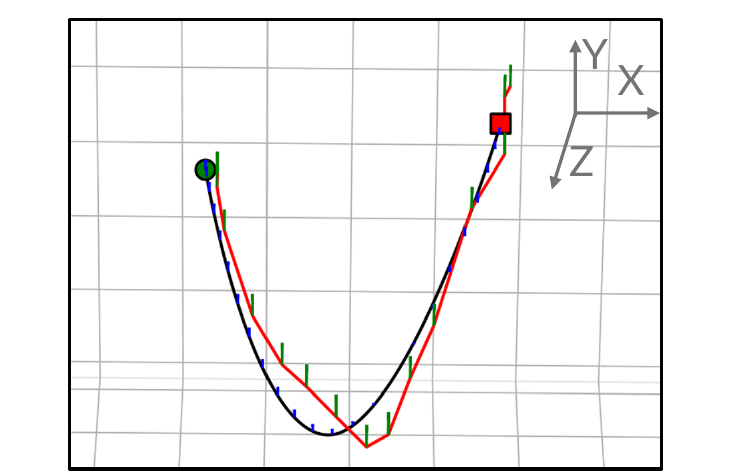}}%\hspace{5pt}
		\subfigure[Scene \#06 (Test) - Mavic3]{\includegraphics[width=0.5\columnwidth]{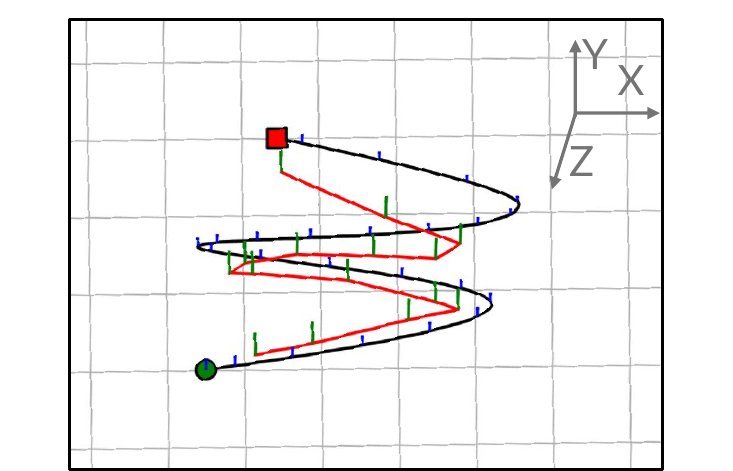}} %\hspace{5pt}
        \subfigure[Scene \#07 (Test) - Air2]{\includegraphics[width=0.5\columnwidth]{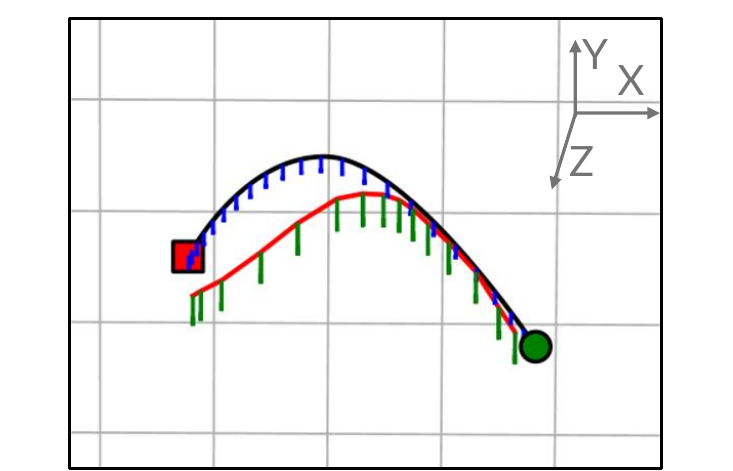}} \vspace{-5pt}
	\caption{\textbf{Qualitative results of the drone 3D pose estimation based on the proposed method.} 
    The results show pose estimation across different validation and test scenes and drone types after applying Gaussian smoothing post-processing.}
	\label{fig:04} \vspace{-5pt}
    \end{figure*}
    % □□□□□□□□□□□□□□□□□□□□□□□□□□□□□□□□□□□□□□□□□□□□□□□□□□□□□□□□□□□□□□□□□□□□□□□□□□□□□
    \subsection{Performance Comparison of Drone 3D Pose Estimation}
    % \textcolor{red}{여기도 다시 봐야됨!!}
        % ㅁ (sb) 본 연구에서는 제안한 DroneKey++의 성능을 기존 드론 자세 추정 방법들과 비교 평가하였다.  
                % ㅁ (sb) 실험은 \texttt{6DroneSyn} 데이터셋의 시나리오 06과 07에서 수행되었으며, 각 시나리오에는 7개 드론 모델에 대한 서로 다른 유형의 시퀀스가 포함되어 총 14개의 테스트 시퀀스로 구성된다.  
                      % ✅ (yjcho) 앞서 설명하였으므로 생략
            %, which contain 14 test sequences across 7 drone models.  
        % ㅁ (sb) 모든 비교 실험은 동일한 합성 데이터셋에서 수행되어 공정한 성능 비교를 보장하였다.
            %All comparison experiments were conducted on the same synthetic dataset to ensure fair performance evaluation.    
            We evaluate the performance of the proposed DroneKey\textcolor{encoderpink}{+}\textcolor{decoderblue}{+} against existing drone pose estimation methods.  
            Experiments were conducted on test scenes \#06 and \#07 of the \texttt{6DroneSyn} dataset.
            Since prior research on drone 3D pose estimation is relatively limited, we designed the comparisons to be as fair and comprehensive as possible. Specifically, we considered the following categories of baselines:
            
            % \noindent $\bullet$ \textbf{Object Detector + PnP}: Keypoints are first detected using a generic detector~\cite{yolov8}, and the 3D pose is then estimated by applying a PnP solver~\cite{pnp} to the centers of the boxes.
            
            \noindent $\bullet$ \textbf{Keypoint Detector + PnP}: Although originally developed for common objects such as humans or vehicles rather than drones, existing keypoint detection~\cite{yolov8} methods were combined with a PnP solver~\cite{pnp} to infer 3D drone poses.
            
            \noindent $\bullet$ \textbf{Drone-specific approaches}: We also compare against drone-focused 3D pose estimation methods, such as DroneKey~\cite{DroneKey} and DronePose~\cite{DronePose}\footnote{For DronePose~\cite{DronePose}, the original 3D mesh-based prior could not be reproduced due to unavailable mesh and dataset. Instead, we reimplemented it using MSE loss on rotation and translation, resulting in a prior-free setting consistent with our evaluation.}, which share a similar goal to ours.

        % ㅁ (sb) 비교에 사용된 기존 방법들에는 몇 가지 제약이 존재하였다.  
            % Several constraints existed when comparing with existing methods.  
        % ㅁ (sb) 예를 들어, Albanis et al.의 DronePose는 드론의 3D 메시를 활용한 실루엣 기반 감독과 복잡한 손실 파이프라인을 요구한다.  
            % For example, DronePose~\cite{DronePose} relies on silhouette-based supervision using 3D meshes and a complex loss pipeline.  
        % ㅁ (sb) 그러나 기종별 정확한 3D mesh를 수집·정합하는 것은 현실적으로 어렵고, 실제 환경에서도 mesh 확보는 거의 불가능하다.  
            % However, collecting and aligning accurate 3D meshes for each drone type is highly impractical, and such meshes are rarely available in real-world scenarios.  
        % ㅁ (sb) 따라서 본 연구에서는 mesh를 사용하지 않는 것이 오히려 현실적이며, 이를 반영하기 위해 mesh 기반 파이프라인을 제외하고 네트워크만 \texttt{6DroneSyn} 데이터셋으로 재학습하여 비교하였다.  
            % Therefore, we consider it more appropriate to exclude mesh supervision, and thus retrained only the network on the \texttt{6DroneSyn} dataset for a fair and realistic comparison.  
        % ㅁ (sb) 한편, Jin et al.의 DronePoseGraph는 공개된 코드가 없어 직접적인 비교가 불가능하였다.  
            %Jin's method~\cite{DronePoseGraph} could not be compared directly due to unavailable public code.  
        % ㅁ (sb) 또한, DroneKey는 PnP solver가 scale ambiguity를 해소하기 위해 각 기종의 real-size가 반드시 필요하므로, 본 연구에서는 원 저자 설정을 따라 real-size 정보를 제공하였다.  
            % DroneKey~\cite{DroneKey} inherently requires the real size of each drone to resolve scale ambiguity in the PnP solver, and we followed the original setting by providing the real-size specifications.  

        \textls[-5]{
        % ㅁ (sb) 정량적 결과는 표 \ref{tab:03}에 요약되어 있다.  
            Quantitative results are summarized in Table \textcolor{skyblue}{\ref{tab:03}}.  
        % ⭐⭐⭐ 비교 공정성 이슈
        % ㅁ (sb) DronePose는 rotation에서 큰 오차를 보였지만 translation은 일부 시퀀스에서 경쟁력이 있었다.  
            Since DronePose~\cite{DronePose} was re-implemented in a prior-free setting, it shows large rotation errors, while translation can be relatively competitive in some sequences.
        % ㅁ (sb) 이는 원래 방법이 3D mesh로 생성한 silhouette mask를 rotation loss에 사용하기 때문에, 본 실험에서 해당 손실 항을 제외하자 rotation 감독이 약화된 결과다.  
            %This is because the original method supervises rotation with silhouette masks rendered from 3D meshes, and excluding this loss in our experiments weakened rotation supervision.  
        % ㅁ (sb) translation 학습은 주로 투영 좌표 일관성(projection consistency)에 의해 제약되므로, rotation loss 제거의 영향이 제한적이다.  
            %Translation learning is primarily constrained by projection consistency in image coordinates, making the effect of removing the rotation loss limited.  
        % ㅁ (sb) DroneKey는 rotation 정확도에서는 개선을 보였지만 translation은 불안정하였다.  
            On the other hand, DroneKey~\cite{DroneKey} achieves high rotation accuracy but exhibits unstable translation. This is because PnP solvers inherently suffer from scale ambiguity and require the actual physical size of each drone to resolve absolute translation.
        % ㅁ (sb) 이는 PnP solver가 본질적으로 scale ambiguity를 가지기 때문에, 절대 거리 추정을 위해 각 드론의 실제 길이(real length)가 반드시 필요하기 때문이다.  
        % ◆ (sb) yolov8n-pose.pt model, 100 epoch
            The Keypoint Detector~\cite{yolov8} + PnP~\cite{pnp} baseline shows competitive $\text{MedAE}_{\textbf{t}}$ errors but suffers from severe outliers and high variance due to frequent keypoint detection failures on small drone targets.
        }

        % ㅁ (sb) 제안한 DroneKey++는 rotation에서 평균 MAE 17.34$^\circ$와 MedAE 17.1$^\circ$를 기록하며, DronePose 대비 78\% 향상, DroneKey 대비 14\% 향상을 달성하였다.  
        % ㅁ (sb) translation에서도 평균 MAE 0.135m와 MedAE 0.242m를 기록하며, DronePose 대비 17\% 향상, DroneKey 대비 75\% 향상을 보였다.  
            Our DroneKey{\color{encoderpink}+}{\color{decoderblue}+} achieves the lowest rotation errors, with MAE of 17.34$^\circ$ and MedAE of 17.1$^\circ$, representing improvements of 78\% over DronePose and 14\% over DroneKey.
            For translation, it yields MAE of 0.135 m and MedAE of 0.242 m, achieving improvements of 17\% over DronePose and 75\% over DroneKey.
        % ㅁ (sb) 이는 class embedding이 드론의 크기와 형상 단서를 내부적으로 학습하고, ray feature가 깊이 정보를 보강하여 두 지표 모두에서 일관된 성능을 달성하기 때문이다.  
            This is because the class embeddings implicitly encode size and shape cues, while the ray features enhance pose estimation robustness, yielding consistent performance across both metrics.

            % ■ (sb) 정성적 결과는 그림 \ref{fig:NN}에서 확인할 수 있으며, 제안한 방법이 다양한 시퀀스에서 보다 안정적이고 정확한 3D pose를 산출함을 보여준다.  
            % ■ (sb) 특히 복잡한 배경이나 다양한 드론 형태에서도 robust하게 동작하는 것을 확인하였다.  
            % ■ (sb) 추가적인 시뮬레이션 결과는 부록에 포함되어 있으며, 실제 드론 비디오에서의 정성적 검증을 통해 제안한 방법의 일반화 능력을 확인했다.
            % ■ (sb) 이는 제안한 DroneKey++가 단순히 정량적 지표에서만 우수할 뿐 아니라 다양한 시나리오에서도 일관된 성능을 보임을 입증한다.  
            Qualitative results in Fig.~\textcolor{skyblue}{\ref{fig:04}} demonstrate that our method provides more stable and accurate 3D poses across diverse sequences.
            The method works robustly even with cluttered backgrounds and varying drone types.
            Additional qualitative results are provided in the supplementary video, including validation on real drone scenarios to verify the generalization capability.
            These results indicate that DroneKey\textcolor{encoderpink}{+}\textcolor{decoderblue}{+} not only outperforms existing methods in quantitative metrics but also delivers consistent performance across diverse scenarios.

    % □□□□□□□□□□□□□□□□□□□□□□□□□□□□□□□□□□□□□□□□□□□□□□□□□□□□□□□□□□□□□□□□□□□□□□□□□□□□□
    % □□□□□□□□□□□□□□□□□□□□□□□□□□□□□□□□□□□□□□□□□□□□□□□□□□□□□□□□□□□□□□□□□□□□□□□□□□□□□
    \subsection{Ablation Studies}
    \label{exp:ablation}

        \subsubsection{\textbf{Keypoint encoder}}
        \textls[-5]{
            % ㅁ (sb) Table\textcolor{skyblue}{\ref{tab:ablation_encoder}}는 2D keypoint와 class label을 동시에 예측하는 keypoint encoder의 효과를 보여준다. 
                Table \textcolor{skyblue}{\ref{tab:ablation_encoder}} presents the effect of the proposed keypoint encoder, which jointly predicts 2D keypoints and class labels.
            % ㅁ (sb) Encoder를 사용하지 않고 ground-truth 주석을 직접 제공한 경우, 회전 추정에서 큰 오차가 발생하였다. 
                When ground-truth annotations are directly provided without the encoder, rotation estimation suffers from large errors. 
            % ㅁ (sb) 반면 encoder를 활성화하면 회전 MAE와 MedAE가 각각 59.12$^\circ$→19.37$^\circ$, 44.28$^\circ$→19.24$^\circ$로 크게 감소하여, end-to-end 예측과 맥락적 특징 전파가 강건한 pose 추정에 핵심적임을 입증한다. 
                By contrast, enabling the encoder significantly reduces both MAE and MedAE in rotation (59.12$^\circ\rightarrow$17.34$^\circ$ and 44.28$^\circ\rightarrow$17.21$^\circ$, respectively), demonstrating that end-to-end prediction and contextual feature propagation are crucial for robust 3D pose estimation.
            % ㅁ (sb) Translation 성능 또한 MAE가 0.159m→0.135m로 개선되어, keypoint encoder가 회전뿐만 아니라 translation 추정에도 도움이 됨을 보여준다. 
                Translation performance is also improved, with MAE reduced from 0.159m to 0.135m, indicating that the keypoint encoder benefits both rotation and translation estimation.
            % ㅁ (sb) 이러한 결과는 외부에서 제공된 라벨을 단순히 활용하거나 keypoint 검출을 분리하여 수행하는 것보다 end-to-end 방식으로 통합 학습하는 접근이 더 효과적임을 입증한다.
            % ✅ (yjcho) encoder 효과 이야기 하다 갑자기 end-to-end (decoder까지) 이야기 하는것은 별로...
            %These results demonstrate that end-to-end joint learning is more effective than using externally provided labels or performing keypoint detection as a separate module.
        }

        % □□□□□□□□□□□□□□□□□□□□□□□□□□□□□□□□□□□□□□□□□□□□□□□□□□□□□□□□□□□□□□□□□□□□□□□□□□□□□
        \subsubsection{\textbf{3D pose decoder}}
        \label{sec:exp:decoder}
        \textls[-2]{
        As illustrated in Fig.~\textcolor{skyblue}{\ref{fig:02}}b, the proposed 3D pose decoder consists of two MLPs ($\text{MLP}_{pose}$, $\text{MLP}_{pose}$) and two embedding layers (ray embedding and class embedding).
        To analyze the contribution of each component, we conducted an ablation study on four decoder configurations:
        \noindent \textbf{(1)} Only $\text{MLP}_{pose}$: directly regresses 3D pose from encoder-predicted 2D keypoints; \textbf{(2)} RayEmbed + $\text{MLP}_{pose}$: converts 2D keypoints into rays using the camera intrinsic matrix, which are then fed into $\text{MLP}_{pose}$; \textbf{(3)} RayEmbed + $\text{MLP}_{pose}$ + $\text{MLP}_{3D}$: adds $\text{MLP}_{3D}$ to infer 3D keypoints before final pose estimation; \textbf{(4)} Full decoder (Ours): integrates all components, including the class embedding, for end-to-end pose estimation.

            \begin{table}[t]
            \vspace{-5pt}
            \setlength{\tabcolsep}{6pt} % 기본은 6pt
            \centering
            \caption{\textbf{Effect of the keypoint encoder.} 
                The encoder jointly predicts 2D keypoints and class labels in an end-to-end manner (\checkmark), 
                while the baseline directly uses ground-truth annotations (blank).}
            \label{tab:ablation_encoder}
            \begin{tabular}{c|cccc}
            \hline \noalign{\hrule height 0.5pt}
                                                        & \multicolumn{2}{c}{$\mathbf{R}$ $(^\circ)$} & \multicolumn{2}{c}{$\mathbf{t}$ (m)} \\
            \multirow{-2}{*}{\textbf{Encoder}} & \textbf{MAE}   & \textbf{MedAE}  & \textbf{MAE}   & \textbf{MedAE}  \\ \hline\noalign{\hrule height 0.5pt}
                                                        &    59.12        &   44.28        &   0.159        &    0.254             \\
            \rowcolor[HTML]{EFEFEF} 
                                         \checkmark      &\textbf{17.34} \scalebox{0.7}{\textcolor{blue}{(-41.78)}}&\textbf{17.10} \scalebox{0.7}{\textcolor{blue}{(-27.18)}}&\textbf{0.135} \scalebox{0.7}{\textcolor{blue}{(-0.024)}} &\textbf{0.242} \scalebox{0.7}{\textcolor{blue}{(-0.012)}} \\ \hline\noalign{\hrule height 0.5pt}
            \end{tabular}
            \end{table}

            \begin{table}[t]
            \setlength{\tabcolsep}{3.5pt} % 기본은 6pt
            \centering
            \caption{
            \textls[-1]{\textbf{Ablation study of the 3D pose decoder with different component configurations.}}}
            %Effect of adding components step by step: 2D keypoint detection, ray transformation, 3D reconstruction, and class feature incorporation.
            \label{tab:ablation_decoder}
            \begin{tabular}{l|cccc}
            \hline \noalign{\hrule height 0.5pt}
                                                                   & \multicolumn{2}{c}{$\mathbf{R}$ $(^\circ)$} & \multicolumn{2}{c}{$\mathbf{t}$ (m)} \\
            \multirow{-2}{*}{\textbf{Decoder Settings}}            & \textbf{MAE}   & \textbf{MedAE}  & \textbf{MAE}   & \textbf{MedAE}   \\ \hline \noalign{\hrule height 0.5pt}
            (1) Only $\text{MLP}_{pose}$                           & 17.40          & 17.19           & 0.219          & 0.372            \\
            (2) RayEmbed + $\text{MLP}_{pose}$                     & 17.76          & 17.86           & 0.153          & 0.248            \\
            (3) RayEmbed + $\text{MLP}_{pose}$ + $\text{MLP}_{3D}$ & 19.71          & 19.43           & 0.158          & 0.296            \\
            \rowcolor[HTML]{EFEFEF} (4) Full decoder (Ours)        & \textbf{17.34} & \textbf{17.10}  & \textbf{0.135} & \textbf{0.242}   \\ \hline \noalign{\hrule height 0.5pt}
            \end{tabular} \vspace{-10pt}
            \end{table}

            \begin{table}[t]
            \vspace{-5pt}
            \setlength{\tabcolsep}{10pt} % 기본은 6pt
                    \centering
                    \caption{\textbf{Comparison of training strategies with different loss weighting schemes.}}
                    \label{tab:06}
            \begin{tabular}{l|cccc}
            \hline \noalign{\hrule height 0.5pt}
                                                     & \multicolumn{2}{c}{$\mathbf{R}$ $(^\circ)$}            & \multicolumn{2}{c}{$\mathbf{t}$ (m)}            \\
            \multirow{-2}{*}{\textbf{Training settings}} & \textbf{MAE}         & \textbf{MedAE}       & \textbf{MAE}         & \textbf{MedAE}       \\ \hline \noalign{\hrule height 0.5pt}
            Tahn-weighted           & 27.30  & 20.40 & 0.192 & 0.303 \\
            Smoothly-shifted        & 19.37 & 19.24 & 0.154 & 0.269 \\
            3D-biased               & 18.38 & 18.62 & 0.165 & 0.293 \\
            \rowcolor[HTML]{EFEFEF} 
            Equal Loss                  & \textbf{17.34} & \textbf{17.10}  & \textbf{0.135} & \textbf{0.242} \\
             \hline \noalign{\hrule height 0.5pt}
            \end{tabular} \vspace{-5pt}
            \end{table}

        The results in Table~\ref{tab:ablation_decoder} highlight the contribution of each decoder component.
        When using only $\text{MLP}_{pose}$, the model achieves reasonable rotation accuracy (17.4$^\circ$) but suffers from a large translation error (0.219m), since 2D observations alone cannot resolve the inherent scale ambiguity of monocular vision.
        Incorporating ray embedding provides geometric depth cues by projecting 2D keypoints into camera rays, thereby improving translation performance to 0.153m. 
        Nevertheless, the absence of absolute scale information leaves the ambiguity unresolved.
        Adding $\text{MLP}_{3D}$ introduces an intermediate step of estimating 3D keypoints before pose regression. 
        
        However, this additional step degrades performance across both rotation and translation, likely because the intermediate estimation introduces accumulated errors rather than providing useful structural cues.
        Finally, the full decoder (ours) integrates class embedding, yielding a translation error of 0.135m and achieving the best overall performance. 
        This demonstrates that class embeddings implicitly encode physical size and shape cues of drone models, enabling the network to infer scale automatically without requiring external priors. As a result, the full decoder achieves the best overall performance across both rotation and translation metrics.
        }

        % □□□□□□□□□□□□□□□□□□□□□□□□□□□□□□□□□□□□□□□□□□□□□□□□□□□□□□□□□□□□□□□□□□□□□□□□□□□□□
        \subsubsection{\textbf{Training strategy}}
        \label{sec:exp:loss}
    \textls[-10]{
            % ㅁ (sb) 표 \ref{tab:04}는 다양한 loss 가중치 전략을 비교한 결과를 보여준다.  
            % ㅁ (sb) Tahn-weighted loss는 학습 초반 절반까지는 2D keypoint 손실에 가중치를 주고, 이후에는 3D 손실에만 집중한다.  
             % ㅁ (sb) Smoothly-shifted loss는 같은 아이디어를 따르되, 전환을 갑작스럽게 하지 않고 점진적으로 진행한다.
             % ㅁ (sb) 3D-biased loss는 3D 항에 큰 비중(예: ×5)을 두어 강하게 강조한다.  
             % ㅁ (sb) 이러한 설계들은 3D 감독을 다른 단계나 크기로 우선시하여 3D 추정을 개선하려는 목적을 가진다.  
               Table~\textcolor{skyblue}{\ref{tab:06}} compares different loss weighting strategies for the encoder loss $\mathcal{L}_{\text{enc}}$ and the decoder loss $\mathcal{L}_{\text{dec}}$.
               The tahn-weighted loss assigns weight to the encoder loss during the first half of training and then shifts entirely to the decoder loss.  
               The tanh-weighted scheme emphasizes the encoder loss during the first half of training and gradually shifts the weight to the decoder loss.
               In addition to the tanh-weighted scheme, the smoothly-shifted loss follows a similar idea but transitions gradually rather than abruptly. Meanwhile, the 3D-biased loss places stronger emphasis on the 3D terms (e.g., assigning a weight five times larger to the 3D pose decoder).
               We evaluated these strategies to better understand how loss weighting influences the balance between encoder and decoder.
               %These designs aim to improve 3D estimation by prioritizing 3D supervision at different stages or magnitudes.  
    
            % ㅁ (sb) 그러나 결과는 이러한 가중치 전략들이 회전과 이동 모두에서 더 높은 오차를 초래함을 보여준다.  
               % However, the results show that these weighting schemes lead to higher errors in both rotation and translation.  
               % In contrast, the Equal Loss strategy, where all loss terms are weighted equally, achieves the best performance in all metrics.
               % A balanced contribution from both 2D and 3D objectives provides more stable optimization, demonstrating that equal weighting is the most effective training strategy for our architecture. 
               % These findings highlight the importance of balancing 2D and 3D supervision for robust 3D pose estimation, showing that balanced multi-task learning is more effective than overemphasizing a single loss. 
               However, the results show that the proposed weighting schemes lead to higher errors in both rotation and translation.
               In contrast, the Equal Loss strategy, where all loss terms are weighted equally, achieves the best performance across all metrics.
               This indicates that enforcing complex weight schedules can destabilize optimization, whereas a simple equal weighting naturally balances the contributions of 2D and 3D objectives.
               These findings highlight that balanced multi-task supervision is more effective than overemphasizing a single objective, making equal weighting the most suitable strategy for our architecture.
            % ㅁ (sb) 반면 모든 손실 항에 동일한 가중치를 부여한 Equal Loss 전략은 모든 지표에서 가장 좋은 성능을 보였다.  
               
            % % ㅁ (sb) 이는 정확한 3D 회전 추정에도 여전히 2D keypoint 손실의 지도가 필요함을 의미한다.  
            %    This indicates that accurate 3D rotation estimation still requires consistent guidance from 2D keypoint losses.  
            % ㅁ (sb) 2D와 3D 목표의 균형 있는 기여가 더 안정적인 최적화를 제공하며, Equal Loss가 본 아키텍처에 가장 효과적인 학습 전략임을 보여준다.  
                
            % ㅁ (sb) 이러한 결과는 3D pose 추정에서 2D와 3D 학습 신호의 균형이 중요하며, 특정 손실을 과도하게 강조하기보다 균형 잡힌 멀티태스크 학습이 더 효과적임을 시사한다.  
                
    }

            \begin{figure}[t]
            % \vspace{-5pt}
        	\centering
                \subfigure[PCA]{\includegraphics[width=0.48\columnwidth]{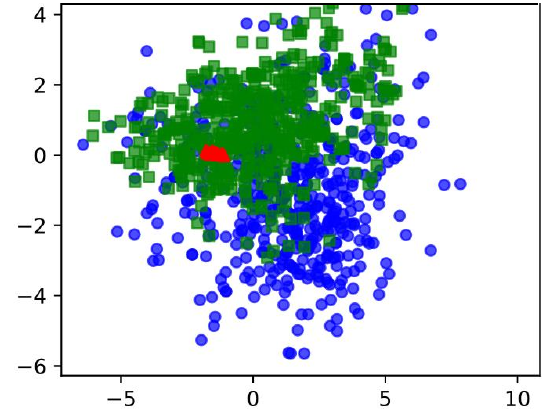}}\hspace{4pt}
        		\subfigure[t-SNE]{\includegraphics[width=0.48\columnwidth]{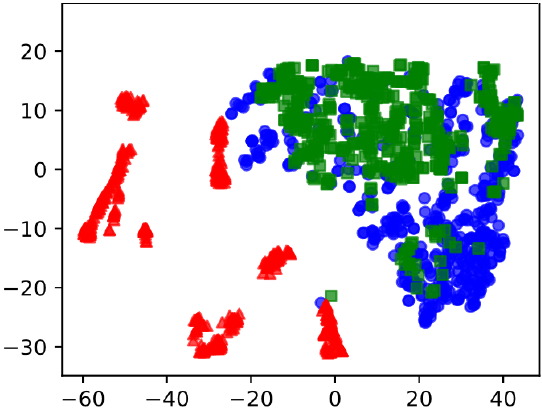}}\vspace{-5pt}
        	\caption{\textbf{Comparison of feature distributions across drone image datasets.} Dimensionality reduction visualization comparing real-world drone images (\textcolor{blue}{●}), existing synthetic dataset (\textcolor{red}{▲}), and our 360-camera-based synthetic dataset (\textcolor{darkgreen}{■}) using (a) PCA with the first two principal components and (b) t-SNE with 2D embedding components.} \vspace{-10pt}
        	\label{fig:06}
        \end{figure}
        
    % □□□□□□□□□□□□□□□□□□□□□□□□□□□□□□□□□□□□□□□□□□□□□□□□□□□□□□□□□□□□□□□□□□□□□□□□□□□□□
    % □□□□□□□□□□□□□□□□□□□□□□□□□□□□□□□□□□□□□□□□□□□□□□□□□□□□□□□□□□□□□□□□□□□□□□□□□□□□□

     \subsection{Evaluating Realism and Diversity of Our Dataset}
     \label{sec:exp:dataset}
        % ㅁ (sb) 합성 데이터의 실제 유사성을 검증하기 위해 세 가지 데이터셋을 비교하였다: (1) 웹에서 수집한 실제 드론 이미지, (2) 3D 모델과 가상 배경으로 생성한 기존 합성 이미지, (3) 360도 배경 이미지와 드론 3D 모델을 합성하여 생성한 우리의 합성 이미지. 각 데이터셋에서 500개의 샘플을 랜덤하게 선택하여 공정한 비교를 보장하였다.  
            % To evaluate the similarity between real and synthetic data, we compared three datasets: 
            To evaluate the realism and diversity of our dataset, we compared it against two baselines:
            \begin{itemize}
                \item Real drone image frames collected from the web.
                \item Existing synthetic 2D drone dataset generated with 3D drone models and virtual backgrounds~\textcolor{skyblue}{\cite{single-drone-traj}}.
                \item Our dataset created by compositing 3D drone models with 360-degree background panoramas.
            \end{itemize}
            For a fair comparison, 500 samples were randomly selected from each dataset.
        % ㅁ (sb) 우리는 두 가지 차원 축소 기법을 적용하였다. PCA(주성분 분석)는 고차원 feature를 주요 변동성을 보존하며 저차원으로 투영하여 데이터의 전체 분포를 시각화할 수 있다. 반면 t-SNE는 국소적 유사성을 보존하여 클러스터 구조와 도메인 차이를 더욱 선명하게 드러낸다. 모든 feature는 StandardScaler로 평균 0, 분산 1로 표준화한 뒤 분석하였다.  
        \textls[-2]{
            We then applied two dimensionality reduction techniques for visualization. Principal component analysis (PCA)~\cite{pca} projects high-dimensional features into a lower-dimensional space while preserving major variance, thereby revealing overall distribution patterns. In contrast, t-distributed stochastic neighbor embedding (t-SNE)~\cite{tsne} preserves local similarity, highlighting cluster structures and domain differences. Before analysis, all features were standardized to zero mean and unit variance using a standard scaler~\cite{sklearn}.
            }
        % ㅁ (sb) 분석 결과(Fig.\ref{fig:06})에서, 기존 합성 데이터(빨간색)는 제한된 영역에 집중되어 다양성이 부족함을 보여준다. 반면 실제 데이터(파란색)와 우리의 합성 데이터(녹색)는 넓게 분포하며, 특히 t-SNE 공간에서 실제 데이터와 우리의 데이터가 중첩되는 것은 360도 카메라 기반 합성의 효과를 입증한다.  

            {\textls[-2]{%
            As shown in Fig.~\textcolor{skyblue}{\ref{fig:06}}, existing synthetic data (\textcolor{red}{▲}) is concentrated in a narrow region, indicating limited diversity.        
            In contrast, real data (\textcolor{blue}{●}) and our synthetic data (\textcolor{darkgreen}{■}) are broadly distributed. 
            In the t-SNE space, our synthetic data overlaps with real data clusters, highlighting the realism and diversity achieved by our 360-degree camera-based synthesis strategy.
            }}

%■■■■■■■■■■■■■■■■■■■■■■■■■■■■■■■■■■■■■■■■■■■■■■■■■■■■■■■■■■■■■■■■■■■■■■■■■■■

\section{Conclusions and Future Works}
\label{sec:conclusions}
\textls[0]{}
% ✨ Conclusions 내용
    % ■ (sb) 본 연구에서는 기존 방법들의 사전 정보 의존성을 해결하는 DroneKey++를 제안하였다.
    % ■ (sb) 키포인트 검출, 드론 분류, 3D 자세 추정을 완전히 통합한 end-to-end 학습 프레임워크를 구현했다.
    % ■ (sb) 클래스 임베딩과 레이 기반 기하학적 추론을 통해 크기 정보 없이도 정확한 3D 자세를 추정한다.
    % ■ (sb) 회전 MAE 17.34$^\circ$, 위치 MAE 0.135m를 달성하며 기존 방법들 대비 크게 개선했고, GPU 414.07 FPS, CPU 19.25 FPS의 실시간 성능을 보였다.
    % ■ (sb) 7개 드론 모델과 88개 배경을 포함한 52,920장의 6DroneSyn 데이터셋을 구축하여 360도 파노라마 기반 합성으로 도메인 갭을 최소화한 고품질 훈련 데이터를 생성했다.
        % We proposed DroneKey\textcolor{encoderpink}{+}\textcolor{decoderblue}{+} to eliminate the prior information dependency of existing methods.
        % We implemented an end-to-end learning framework that fully integrates keypoint detection, drone classification, and 3D pose estimation.
        % It estimates accurate 3D poses without size information through class embeddings and ray-based geometric reasoning.
        % We achieved rotation MAE 17.34$^\circ$ and translation MAE 0.135m, significantly improving over existing methods, with real-time performance of 414.07 FPS on GPU and 19.25 FPS on CPU.
        % We built the 6DroneSyn dataset with 52,920 images covering 7 drone models and 88 backgrounds, generating high-quality training data with minimized domain gap through 360-degree panorama-based synthesis.

        In this work, we presented DroneKey\textcolor{encoderpink}{+}\textcolor{decoderblue}{+}, a prior-free end-to-end framework that integrates keypoint detection, drone classification, and 3D pose estimation. By leveraging class embeddings and ray-based geometric reasoning, our method estimates 3D poses accurately without requiring prior size information.
        It achieves rotation MAE 17.34$^\circ$ and translation MAE 0.135 m, significantly outperforming existing methods, while running in real time at 414.07 FPS on GPU and 19.25 FPS on CPU.
        In addition, we introduced the \texttt{6DroneSyn} dataset with 52,920 images across 7 drone models and 88 backgrounds, generated via 360-degree panorama-based synthesis. 
        This dataset provides high-quality annotations and reduced domain gap, offering a valuable benchmark for advancing future research in drone pose estimation.

        This work identifies three main directions for future research.
        First, we will explore strategies to further balance dataset realism and learnability, ensuring both reduced domain gap and effective training.
        Second, we will develop keypoint confidence mechanisms and integrate them into the 3D pose decoder to improve robustness under occlusion.
        Third, we plan to expand our dataset with more diverse drone models and environments, and conduct real-world validation to complete our methodology. We expect that these extensions will further strengthen the utility of our dataset and framework for future drone-related research.

%■■■■■■■■■■■■■■■■■■■■■■■■■■■■■■■■■■■■■■■■■■■■■■■■■■■■■■■■■■■■■■■■■■■■■■■■■■■

% \section*{APPENDIX}
% The code and dataset are available at \url{https://github.com/kkanuseobin/DroneKey}.

%■■■■■■■■■■■■■■■■■■■■■■■■■■■■■■■■■■■■■■■■■■■■■■■■■■■■■■■■■■■■■■■■■■■■■■■■■■■

\section*{ACKNOWLEDGMENT}
% 중견, G5, AX
\linespread{0.85}
\textls[-5]{
\footnotesize
This work was supported by the National Research Foundation of Korea (NRF) grant funded by the Korea government (MSIT) (No. RS-2025-24683045), the Innovative Human Resource Development for Local Intellectualization Program through the Institute of Information \& Communications Technology Planning \& Evaluation (IITP) funded by the Korea government (MSIT) (No. IITP-2024-RS-2022-00156287), and the AX Demonstration Support Program for Manufacturing AI Open Lab funded by the National IT Industry Promotion Agency (NIPA) (No. 0000000).
}
%■■■■■■■■■■■■■■■■■■■■■■■■■■■■■■■■■■■■■■■■■■■■■■■■■■■■■■■■■■■■■■■■■■■■■■■■■■■
\linespread{0.87} % 기본값은 1, 0.9로 줄이면 간격이 좁아짐

\textls[-20]{ % 자간 줄이는 기능

}

\end{document}